\pdfoutput=1
\documentclass[11pt]{article}

\usepackage[letterpaper,margin=1in]{geometry}
\usepackage[T1]{fontenc}
\usepackage{stix2}
\usepackage[scaled=0.88]{beramono}
\usepackage{microtype}
\usepackage{xcolor}
\usepackage{graphicx}
\usepackage{booktabs,multirow,array,tabularx,colortbl}
\usepackage{pifont}
\usepackage{amsmath,mathtools}
\usepackage{caption,subcaption}
\usepackage{enumitem}
\usepackage{titlesec}
\usepackage{fancyhdr}
\usepackage[most]{tcolorbox}
\usepackage{tikz}
\usetikzlibrary{arrows.meta,positioning,calc,shapes.geometric,fit,backgrounds,shadows}
\usepackage{pgfplots}
\pgfplotsset{compat=1.18}
\usepgfplotslibrary{groupplots,fillbetween}
\pgfplotsset{/pgfplots/ybar legend/.style={/pgfplots/legend image code/.code={\draw[##1,/tikz/.cd,yshift=-0.3em] (0cm,0cm) rectangle (0.7em,0.7em);}}}
\usepackage[numbers,sort&compress]{natbib}
\usepackage[colorlinks=true,linkcolor=fxlink,citecolor=fxlink,urlcolor=fxlink]{hyperref}
\usepackage[capitalise,noabbrev]{cleveref}
\hypersetup{pdftitle={Calibrated Decision Rewards: Training Language Models with a Judge That Knows How Sure It Is},pdfauthor={Rudransh Agnihotri, Manasvi Kapoor (FuturixAI)},pdfsubject={Reinforcement learning, reward models, calibration},pdfkeywords={RLHF, GRPO, reward models, LLM-as-a-judge, calibration, decision models}}
\usepackage{fontawesome5}

\definecolor{fxink}{HTML}{0A0B0C}
\definecolor{fxgraphite}{HTML}{303030}
\definecolor{fxsteel}{HTML}{4983B5}
\definecolor{fxstone}{HTML}{E9EAE6}
\definecolor{fxlime}{HTML}{D3F072}
\definecolor{fxslate}{HTML}{62676B}
\definecolor{fxlink}{HTML}{2F6A9C}     %
\definecolor{fxblue}{HTML}{0A0B0C}     %
\definecolor{fxteal}{HTML}{2F6A9C}     %
\definecolor{fxorange}{HTML}{62676B}   %
\definecolor{fxgray}{HTML}{62676B}
\definecolor{fxlight}{HTML}{F4F4F2}
\definecolor{cOurs}{HTML}{4983B5}      %
\definecolor{cSoft}{HTML}{8A8F93}      %
\definecolor{cGT}{HTML}{0A0B0C}        %
\definecolor{cRand}{HTML}{C2C5C3}      %
\definecolor{cBase}{HTML}{DCDCD8}      %
\definecolor{cPRM}{HTML}{303030}       %
\definecolor{cScalar}{HTML}{303030}    %
\definecolor{cHyb}{HTML}{A4C1DA}       %
\definecolor{pending}{HTML}{B03A2E}

\pdfmapline{+fdmsr8t <ec.enc <DMSans-Regular.ttf}
\pdfmapline{+fdmsm8t <ec.enc <DMSans-Medium.ttf}
\DeclareFontFamily{T1}{fdms}{}
\DeclareFontShape{T1}{fdms}{m}{n}{<-> fdmsr8t}{}
\DeclareFontShape{T1}{fdms}{b}{n}{<-> fdmsm8t}{}
\DeclareFontShape{T1}{fdms}{bx}{n}{<-> ssub * fdms/b/n}{}
\DeclareFontShape{T1}{fdms}{m}{it}{<-> ssub * fdms/m/n}{}
\DeclareFontShape{T1}{fdms}{b}{it}{<-> ssub * fdms/b/n}{}
\DeclareFontShape{T1}{fdms}{bx}{it}{<-> ssub * fdms/b/n}{}
\DeclareFontShape{T1}{fdms}{m}{sc}{<-> ssub * fdms/m/n}{}
\newcommand{\brandfont}{\fontfamily{fdms}\fontseries{m}\fontshape{n}\selectfont}
\newcommand{\brandmed}{\fontfamily{fdms}\fontseries{b}\fontshape{n}\selectfont}
\newcommand{\brandlabel}[1]{\textls[80]{\brandmed\MakeUppercase{#1}}}

\titleformat{\section}{\brandmed\Large\color{fxink}}{\textcolor{fxsteel}{\thesection}}{0.7em}{}
\titleformat{\subsection}{\brandmed\large\color{fxink}}{\textcolor{fxsteel}{\thesubsection}}{0.6em}{}
\titleformat{\paragraph}[runin]{\brandmed\normalsize\color{fxink}}{}{0em}{}
\titlespacing*{\section}{0pt}{1.5em}{0.6em}
\titlespacing*{\subsection}{0pt}{1.1em}{0.4em}
\renewcommand{\headrulewidth}{0.5pt}
\renewcommand{\headrule}{\hbox to\headwidth{\color{fxstone}\leaders\hrule height \headrulewidth\hfill}}

\setlist{itemsep=2pt,topsep=3pt,leftmargin=1.3em}
\DeclareCaptionFont{brand}{\brandmed}
\renewenvironment{abstract}{\small\noindent{\color{fxslate}\brandlabel{Abstract}}\par\vspace{3pt}}{\par}

\newcommand{\method}{\textsc{RLCDR}}
\newcommand{\fx}{FxAI-DM}
\newcommand{\up}[1]{\textcolor{fxteal}{\textbf{#1}}}

\newcommand{\best}[1]{\textbf{#1}}
\newcommand{\cmark}{\textcolor{fxteal}{\ding{51}}}
\newcommand{\xmark}{\textcolor{fxorange!80!black}{\ding{55}}}

\newtcolorbox{keybox}[1][]{enhanced,colback=fxlight,colframe=fxink,boxrule=0.5pt,arc=1pt,
  left=8pt,right=8pt,top=6pt,bottom=6pt,fonttitle=\brandmed\small,coltitle=white,
  colbacktitle=fxink,title={#1},
  overlay={\fill[fxlime] (title.south west) rectangle ([yshift=2.2pt]title.south east);}}
\newtcolorbox{qbox}{enhanced,colback=white,colframe=fxgray!50,boxrule=0.4pt,arc=2pt,
  left=6pt,right=6pt,top=4pt,bottom=4pt,fontupper=\small}
\newtcolorbox{pendingbox}{enhanced,colback=pending!5,colframe=pending,boxrule=0.6pt,arc=2pt,
  left=8pt,right=8pt,top=5pt,bottom=5pt,fontupper=\small}

\begin{document}

\thispagestyle{empty}
\noindent\includegraphics[height=17pt]{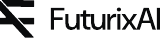}\hfill
{\footnotesize\color{fxslate}\brandlabel{Technical Report}\quad$\cdot$\quad\brandlabel{November 2026}}\par
\vspace{30pt}
\noindent{\color{fxlime}\rule{2.6cm}{4pt}}\par
\vspace{12pt}
\noindent{\brandmed\fontsize{25}{27.5}\selectfont\color{fxink}%
\textls[-20]{Calibrated Decision Rewards:}\\[2pt]
\textls[-20]{Training Language Models with a Judge}\\[2pt]
\textls[-20]{That Knows How Sure It Is}\par}
\vspace{16pt}
\noindent
\begin{minipage}[t]{0.46\linewidth}
{\brandmed\large\color{fxink}Rudransh Agnihotri}\par\vspace{2pt}
{\brandfont\small\href{mailto:rudransh.agnihotri@futurixai.com}{\color{fxslate}rudransh.agnihotri@futurixai.com}}
\end{minipage}%
\begin{minipage}[t]{0.46\linewidth}
{\brandmed\large\color{fxink}Manasvi Kapoor}\par\vspace{2pt}
{\brandfont\small\href{mailto:manasvi.kapoor@futurixai.com}{\color{fxslate}manasvi.kapoor@futurixai.com}}
\end{minipage}\par
\vspace{9pt}
\noindent{\footnotesize\color{fxslate}\brandlabel{FuturixAI Research}\quad$\cdot$\quad\brandlabel{futurixai.com}}\par
\vspace{10pt}
\noindent{\color{fxstone}\rule{\linewidth}{0.8pt}}\par
\vspace{6pt}

\begin{abstract}
\noindent
Reinforcement learning (RL) can only teach a language model what its reward can recognize. Today's rewards force a trade-off. Answer checkers are exact but exist only for tasks with a reference answer. Learned reward models and LLM judges work everywhere, but their scores are not probabilities: they are overconfident, position-biased and easy to game. We study \method{} (\emph{Reinforcement Learning from Calibrated Decision Rewards}). It turns an LLM into a \emph{decision model} that answers a typed question about a response (``is the final answer correct?'', ``which step is the first error?'') with a probability distribution, read in a single forward pass, debiased over option orders and calibrated with about a hundred labels.

\textbf{Objective.} We test one question end to end: \emph{can a calibrated, reference-free decision reward make a real language model better at hard reasoning, as much as the answer key does?} We state six hypotheses in advance and test each with a control designed to falsify it.

\textbf{The policy really improves.} We train Qwen3-1.7B with GRPO, giving every run the same recipe and changing only the reward. With our reward, which \emph{never sees the reference answer}, accuracy on MATH500 rises from \textbf{72.2\% to 76.8\%} (+4.65 points, $p<10^{-4}$), and on the hardest MATH500 level from 44.8\% to 51.7\%. This statistically matches training on the answer key itself (76.9\%). A random-reward control gains only +0.65 (n.s.), so the improvement is caused by the reward, not by RL alone. We also map where the reference-free reward stops being enough. Trained only on problems the base model mostly fails, it still improves the policy (+1.9 MATH500, $p=0.03$) but trails the answer key (74.1\% vs.\ 75.3\%, n.s.), because its probabilities become overconfident on such problems.

The reward stays trustworthy while the policy changes. On-policy, its calibration error is half that of a softmax judge built on the same 27B backbone (ECE 0.074 vs.\ 0.141). It ranks correct above incorrect answers more often at every difficulty level, and shows no sign of reward hacking over training. The same model also locates the first wrong step in a solution better than a dedicated 7B process reward model (+4 to +5 ProcessBench F1 on every split), and the steps it flags in our policy's failed attempts are real errors. We describe the full protocol, including random-reward and same-backbone controls, so that every gain can be attributed.
\end{abstract}

\vspace{2pt}
\begin{keybox}[\faIcon{chart-line}\;\; Key results at a glance]
\small
\begin{tabularx}{\linewidth}{@{}>{\raggedright\arraybackslash}p{0.30\linewidth}X@{}}
\textbf{The model gets better} & MATH500 72.2\% $\rightarrow$ \textbf{76.8\%} (+4.65, $p<10^{-4}$); hardest level 44.8\% $\rightarrow$ 51.7\% ($p=0.002$); pass@4 85.2\% $\rightarrow$ 88.4\%. \\[2pt]
\textbf{Because of the reward} & Random reward: +0.65 (n.s.). \method{} vs.\ random: \textbf{+4.0} ($p=10^{-4}$). \\[2pt]
\textbf{As good as the answer key} & 76.8\% vs.\ 76.9\% with ground-truth reward (difference $-0.1$, 95\% CI $[-1.8, 1.5]$), with no access to reference answers. \\[2pt]
\textbf{Where it stops being enough} & Training only on problems the base model mostly fails, the reference-free reward still helps (MATH500 +1.9, $p=0.03$) but trails the answer key (74.1\% vs.\ 75.3\%, n.s.). Here its step credit works best \emph{added to} the answer key (75.7\%). \\[2pt]
\textbf{A more honest reward} & On-policy ECE \textbf{0.074} vs.\ 0.141 (softmax judge, same backbone); Brier 0.077 vs.\ 0.106; better ranking at every difficulty level. \\[2pt]
\textbf{Finds reasoning errors} & ProcessBench F1 0.772 vs.\ 0.722 (Qwen2.5-Math-PRM-7B) and 0.682 (softmax), winning 20/20 splits. \\[2pt]
\textbf{Cheap} & One forward pass per question; no generated critique, no sampling.
\end{tabularx}
\end{keybox}

\begin{figure}[t]
\centering
\begin{tikzpicture}
\begin{groupplot}[group style={group size=2 by 1,horizontal sep=1.7cm},
  height=5.2cm, tick label style={font=\small}, label style={font=\small},
  title style={font=\small\bfseries,yshift=-2pt}]
\nextgroupplot[width=7.4cm, ybar, bar width=17pt, ymin=70, ymax=80,
  ylabel={MATH500 pass@1 (\%)}, title={After 100 RL steps (ground-truth evaluation)},
  symbolic x coords={Base,Random,Softmax,Ours,Answer key}, xtick={Base,Random,Softmax,Ours,Answer key},
  xticklabels={Base,Random,Softmax\\judge,\textbf{Ours},Answer\\key},
  x tick label style={font=\scriptsize,align=center}, bar shift=0pt,
  error bars/error bar style={black,thick},
  enlarge x limits=0.14, axis x line*=bottom, axis y line*=left, ymajorgrids, grid style={gray!15}]
\addplot+[draw=none,fill=cBase] coordinates {(Base,72.15)};
\addplot+[draw=none,fill=cRand,error bars/.cd,y dir=both,y explicit] coordinates {(Random,72.80) +- (0,1.90)};
\addplot+[draw=none,fill=cSoft,error bars/.cd,y dir=both,y explicit] coordinates {(Softmax,75.50) +- (0,1.80)};
\addplot+[draw=none,fill=cOurs,error bars/.cd,y dir=both,y explicit] coordinates {(Ours,76.80) +- (0,1.85)};
\addplot+[draw=none,fill=cGT,error bars/.cd,y dir=both,y explicit] coordinates {(Answer key,76.90) +- (0,1.78)};
\node[font=\scriptsize\bfseries,text=black,anchor=south] at (axis cs:Base,70.15) {72.2};
\node[font=\scriptsize\bfseries,text=black,anchor=south] at (axis cs:Random,70.15) {72.8};
\node[font=\scriptsize\bfseries,text=white,anchor=south] at (axis cs:Softmax,70.15) {75.5};
\node[font=\scriptsize\bfseries,text=white,anchor=south] at (axis cs:Ours,70.15) {76.8};
\node[font=\scriptsize\bfseries,text=white,anchor=south] at (axis cs:Answer key,70.15) {76.9};
\nextgroupplot[width=7.4cm, xmin=-5, xmax=105, ymin=70.5, ymax=78, xtick={0,25,50,75,100},
  xlabel={RL step}, ylabel={MATH500 pass@1 (\%)}, title={Accuracy during training},
  legend style={font=\scriptsize,at={(0.5,-0.30)},anchor=north,draw=none,legend columns=2,column sep=6pt},
  axis x line*=bottom, axis y line*=left, ymajorgrids, grid style={gray!15}]
\addplot[cOurs,very thick,mark=*,mark size=2.2pt] coordinates {(0,72.15)(25,73.90)(50,73.90)(75,75.60)(100,76.80)};
\addlegendentry{\method{} (ours)}
\addplot[cGT,thick,dashed,mark=square*,mark size=1.8pt] coordinates {(0,72.15)(50,74.40)(100,76.90)};
\addlegendentry{Answer key (ceiling)}
\addplot[cSoft,thick,mark=triangle*,mark size=2.2pt] coordinates {(0,72.15)(50,74.75)(100,75.50)};
\addlegendentry{Softmax judge}
\addplot[cRand,thick,mark=diamond*,mark size=2.2pt] coordinates {(0,72.15)(50,71.20)(100,72.80)};
\addlegendentry{Random reward (floor)}
\end{groupplot}
\end{tikzpicture}
\caption{\textbf{The policy improves because of the reward.} Qwen3-1.7B trained with GRPO on MATH (training split), changing \emph{only} the reward. Every evaluation uses reference answers, never the reward model. \textbf{Left:} accuracy after 100 steps, with 95\% paired-bootstrap intervals for the gain over the base model. \textbf{Right:} our reward improves the policy at every checkpoint and ends level with the answer key. Random reward does not help. Each evaluated checkpoint carries about $\pm1.8$ points of 95\% uncertainty.}
\label{fig:teaser}
\end{figure}
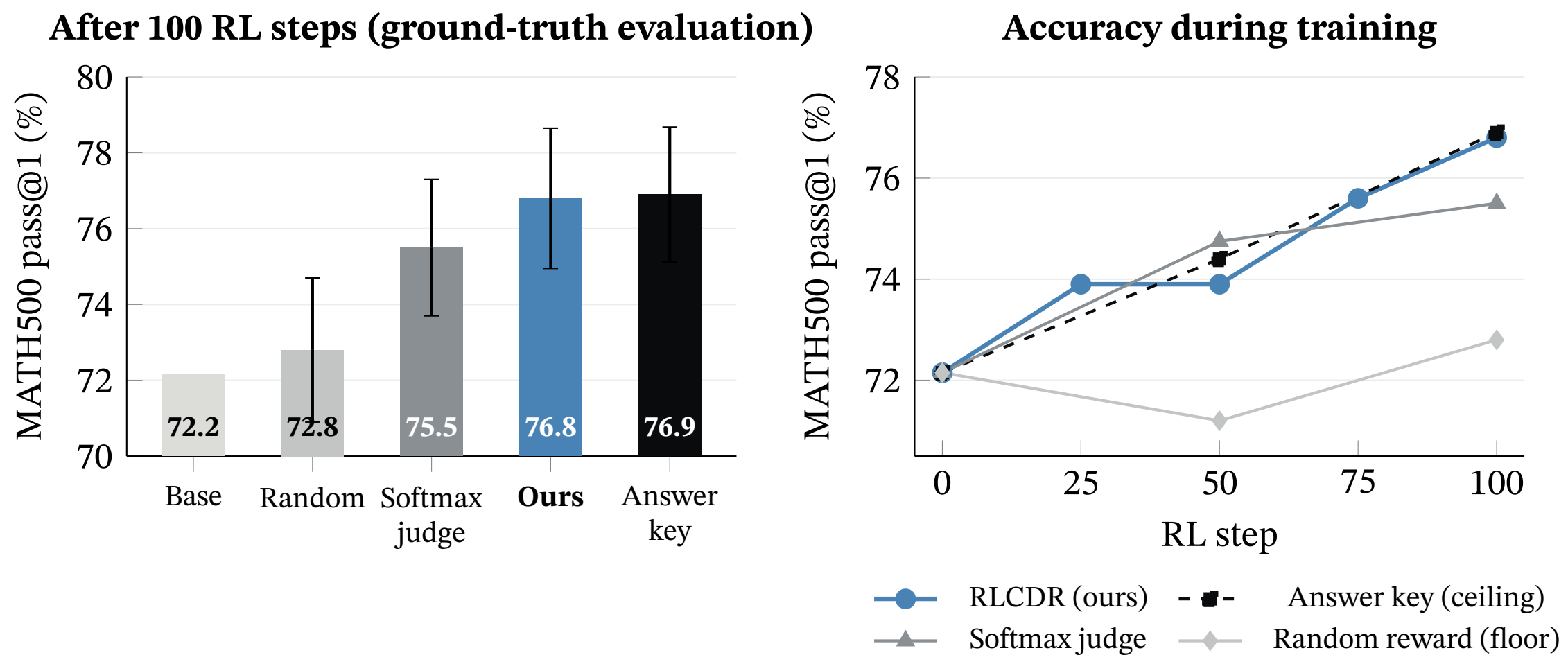

\section{Introduction}
\label{sec:intro}

RL from verifiable rewards has driven much of the recent progress in LLM reasoning \citep{shao2024deepseekmath,deepseek2025r1,lambert2024tulu3,yu2025dapo}. Its strength and its limit are the same thing: an exact answer checker. Where a checker exists (final answers in math, unit tests in code), RL works. Where none exists (proofs, partial credit, open-ended answers, most of alignment), practitioners use learned rewards, and learned rewards are fragile. Bradley--Terry reward models output scores that are only meaningful within one prompt \citep{stiennon2020summarize,ouyang2022instructgpt}. They are overoptimized as training proceeds \citep{gao2022overopt,coste2023ensembles} and pick up spurious features such as length \citep{singhal2023length}. LLM judges read through their next-token softmax are badly overconfident \citep{tian2025overconf,openai2023gpt4,kadavath2022know}, sensitive to option order \citep{zheng2023pride,pezeshkpour2023order,shi2024position} and foolable by single tokens \citep{zhao2025masterkey}. The strongest judges buy accuracy by generating long critiques sampled many times \citep{zhang2024genrm,liu2025grm,whitehouse2025j1}, which makes them a large share of the cost of each RL step \citep{zhou2025ruscarl,ma2025generalreasoner}. DeepSeek-R1 explicitly avoided neural reward models for reasoning because of hacking and cost \citep{deepseek2025r1}.

We argue that the missing ingredient is \emph{calibration}: a reward whose numbers mean what they say. If a reward says ``0.8'', then about 80\% of the responses it scores that way should be correct, and this should keep holding as the policy changes. Calibration is what lets a reward be compared across prompts. It lets the reward express doubt instead of confident error, and it makes the KL-regularized RL objective a principled Bayesian update \citep{korbak2022bayes}.

\begin{keybox}[\faIcon{search}\;\; The research gap]
\small
\begin{enumerate}[label=\textbf{G\arabic*.},leftmargin=2.0em,itemsep=3pt]
  \item \textbf{No reward is both reference-free and trustworthy.} Answer checkers need a reference answer; reward models and judges do not, but their scores are not probabilities and are known to be overconfident, position-biased and hackable (\cref{sec:related}).
  \item \textbf{Calibration of rewards is measured offline, if at all.} Prior work reports judge calibration on fixed benchmarks. Nobody has measured whether a reward stays calibrated \emph{on the policy's own, shifting samples during RL}, which is when it matters.
  \item \textbf{Reward quality is rarely tied to model quality.} Reward benchmarks predict RL outcomes poorly \citep{wen2024rethinking}, and RL gains on Qwen models can appear with random rewards \citep{shao2025spurious}. Few papers show, with a random-reward floor and an answer-key ceiling, that \emph{their} reward is what improved the model.
  \item \textbf{Hard problems give a 0/1 checker nothing to learn from.} When all sampled answers fail, the answer key returns all zeros and GRPO receives no gradient. A reward that can grade \emph{how far} a failed attempt got has not been tested as a calibrated, single-pass signal.
\end{enumerate}
\end{keybox}

\paragraph{Objective.} Our objective is to show that an LLM converted into a \emph{decision model}, debiased and calibrated with about a hundred labels, is a reward that (i)~gives better probabilities than the same LLM read through its softmax, (ii)~stays calibrated on-policy, (iii)~rewards reasoning steps as well as final answers, and above all (iv)~\emph{makes a real model more accurate}, as much as the answer key does, without seeing a reference answer.

\paragraph{Our approach.} We use a \emph{decision model}: an LLM converted, via a trained readout head, into a model that answers a typed multiple-choice question with a probability distribution over the options. The reward is that distribution, after (i)~averaging over option orders to remove position bias and (ii)~fitting two calibration parameters on a small labeled set (\cref{sec:method}). A single forward pass gives an outcome reward (``is the final answer correct?'') or a step-level reward (``which step is the first error?'').

\paragraph{Research questions and hypotheses.} We fixed six hypotheses before running the RL experiments. \Cref{tab:hyp} lists each one, the experiment that tests it, the control that could falsify it, and the verdict. \Cref{sec:verdicts} returns to this table with the evidence.

\begin{table}[h]
\centering\small
\caption{\textbf{Hypotheses, how each is tested, and the outcome.} \textcolor{fxteal}{\textbf{Supported}}: significant in the predicted direction. \textcolor{fxorange}{\textbf{Partly}}: the predicted direction, not yet significant, or significant only on some measures.}
\label{tab:hyp}
\begin{tabularx}{\linewidth}{@{}p{0.55cm}X >{\raggedright\arraybackslash}p{3.6cm} >{\raggedright\arraybackslash}p{2.0cm}@{}}
\toprule
& Hypothesis & Test (falsifying control) & Verdict \\
\midrule
\textbf{H1} & RL with the calibrated decision reward makes the policy more accurate on held-out problems, and the reward, not RL itself, is the cause. & E6: GRPO, ground-truth eval (random-reward floor) & \textcolor{fxteal}{\textbf{Supported}} \\
\textbf{H2} & The reference-free reward trains the policy as well as the answer key. & E6 (answer-key ceiling); E6b & \textcolor{fxteal}{\textbf{Supported}} on MATH; \textcolor{fxorange}{\textbf{Partly}} on hard problems \\
\textbf{H3} & The reward stays calibrated and unhacked on the policy's own samples, better than a softmax judge on the same backbone. & E6 on-policy logs (same-backbone softmax) & \textcolor{fxteal}{\textbf{Supported}} \\
\textbf{H4} & The decision model locates reasoning errors at least as well as a dedicated process reward model, with no step training. & E4 ProcessBench; E6b prefix restarts (random step) & \textcolor{fxteal}{\textbf{Supported}} \\
\textbf{H5} & With equal calibration labels, decision probabilities are better reward probabilities for correctness than softmax probabilities and scalar-RM scores. & E1, E2 (same labels, same backbone, head refit) & \textcolor{fxteal}{\textbf{Supported}} vs.\ softmax; \textcolor{fxorange}{\textbf{Partly}} vs.\ scalar RM \\
\textbf{H6} & On hard problems, step-level decision credit gives a better policy than the answer key alone, and a policy better than the softmax judge on the full set. & E6b; E6 (softmax judge) & \textcolor{fxorange}{\textbf{Partly}} \\
\bottomrule
\end{tabularx}
\end{table}

\paragraph{What we show.} This paper makes one central claim and supports it with controls designed to rule out the usual alternative explanations:

\begin{enumerate}[label=\textbf{C\arabic*.}]
  \item \textbf{A real model improves, and the reward is the cause} (\cref{sec:rl}). With every other setting identical, RL with our reference-free reward improves Qwen3-1.7B by +4.65 MATH500 points. That equals training on the answer key. A random-reward control rules out ``RL alone helps'', which is a known confound for Qwen models \citep{shao2025spurious}.
  \item \textbf{The reward stays calibrated under policy shift} (\cref{sec:faithful}). On 25,600 on-policy samples, its calibration error is half that of the same 27B backbone read through softmax. It orders correct above incorrect answers more often in every difficulty bucket, and its reward--truth gap stays stable across training: no hacking.
  \item \textbf{It rewards reasoning, not just answers} (\cref{sec:steps}). Without any step-level training, it beats a dedicated 7B process reward model at locating the first error (ProcessBench, 20/20 splits). On our own policy's failed attempts, re-solving from the step it flags shows that it finds real errors (a random step shows none).
  \item \textbf{Decision probabilities are better reward probabilities than softmax ones, given equal labels} (\cref{sec:offline}). We compare with proper scoring rules (Brier, log-loss) after giving every method the same calibration labels. The decision readout is better on 3 of 4 reward benchmarks in all 20 splits and reaches its best quality with about 100 labels.
  \item \textbf{We map where the reference-free reward stops being enough} (\cref{sec:hard}). On the 1,758 training problems the base model mostly fails, the reference-free reward still improves the policy but trails the answer key. On-policy, its probabilities become overconfident there (mean 0.28 against a true accuracy of 0.17). Its step credit is most useful added to the answer key.
\end{enumerate}

\begin{figure}[t]
\centering
\resizebox{\linewidth}{!}{%
\begin{tikzpicture}[font=\scriptsize,
  ex/.style={draw=#1,fill=#1!8,rounded corners=3pt,thick,align=center,text width=2.75cm,minimum height=1.45cm,inner sep=3pt},
  hy/.style={draw=fxgray,fill=white,rounded corners=2pt,align=center,text width=1.25cm,minimum height=0.6cm,font=\scriptsize\bfseries},
  arr/.style={-{Stealth[length=2mm]},thick,draw=fxgray!80}]
\node[font=\small\brandmed\color{fxslate}] at (1.5,1.25) {1. Is it a good reward?};
\node[font=\small\brandmed\color{fxslate}] at (7.65,1.25) {2. Does a real model improve?};
\node[font=\small\brandmed\color{fxslate}] at (13.0,1.25) {3. Hardest regime};
\node[ex=fxblue,minimum height=2.05cm,anchor=north] (e1) at (0,0.88) {\textbf{E1} Equal-label calibration\\11.4k items, 4 benchmarks\\\textit{decision vs.\ softmax vs.\ scalar RM}};
\node[ex=fxblue,minimum height=2.05cm,anchor=north] (e2) at (3.05,0.88) {\textbf{E2} Head refit\\same labels, base vs.\ decision features\\\textit{(appendix)}};
\node[ex=fxblue,anchor=north] (e4) at (1.5,-1.42) {\textbf{E4} Step rewards\\ProcessBench, 3,400 solutions\\\textit{vs.\ 7B PRM, softmax}};
\node[ex=fxteal,text width=3.5cm,minimum height=3.15cm] (e6) at (7.65,-0.87) {\textbf{E6} GRPO on MATH\\Qwen3-1.7B, 100 steps\\[3pt]5 arms, only the reward differs:\\\method{} $\cdot$ softmax judge\\answer key (ceiling) $\cdot$ random (floor)\\$\cdot$ base\\[3pt]\textit{ground-truth evaluation +\\on-policy reward logging}};
\node[ex=fxorange,text width=3.0cm,minimum height=3.15cm] (e6b) at (13.0,-0.87) {\textbf{E6b} Hard problems\\1,758 problems with base pass $\le$3/8\\[3pt]critic diagnostics,\\prefix restarts\\[3pt]GRPO 50 steps:\\\fx{} only $\cdot$ answer key\\$\cdot$ answer key + step credit};
\draw[arr] (e1.east) -- (e2.west);
\draw[arr] (e2.east) -- ++(0.35,0) |- ([yshift=6mm]e6.west);
\draw[arr] (e4.east) -- ++(1.15,0) |- ([yshift=-6mm]e6.west);
\draw[arr] (e6.east) -- (e6b.west);
\node[hy] (h5a) at (0,-3.45) {H5};
\node[hy] (h4) at (1.5,-3.45) {H4};
\node[hy] (h5b) at (3.05,-3.45) {H5};
\node[hy] (h12) at (6.95,-3.45) {H1, H2};
\node[hy] (h3) at (8.35,-3.45) {H3};
\node[hy] (h46) at (13.0,-3.45) {H4, H6};
\draw[fxgray!60,densely dotted] (e1.south) -- (h5a.north);
\draw[fxgray!60,densely dotted] (e4.south) -- (h4.north);
\draw[fxgray!60,densely dotted] (e2.south) -- (h5b.north);
\draw[fxgray!60,densely dotted] (e6.south -| h12.north) -- (h12.north);
\draw[fxgray!60,densely dotted] (e6.south -| h3.north) -- (h3.north);
\draw[fxgray!60,densely dotted] (e6b.south) -- (h46.north);
\end{tikzpicture}}
\caption{\textbf{The experimental program.} Stage 1 asks whether the decision model is a good reward on fixed data (E1, E2, E4). Stage 2 asks whether training with it improves a real model, with controls that isolate the reward (E6). Stage 3 tests the regime where the answer key gives no signal (E6b). The boxes at the bottom name the hypotheses (\cref{tab:hyp}) that each experiment tests.}
\label{fig:roadmap}
\end{figure}
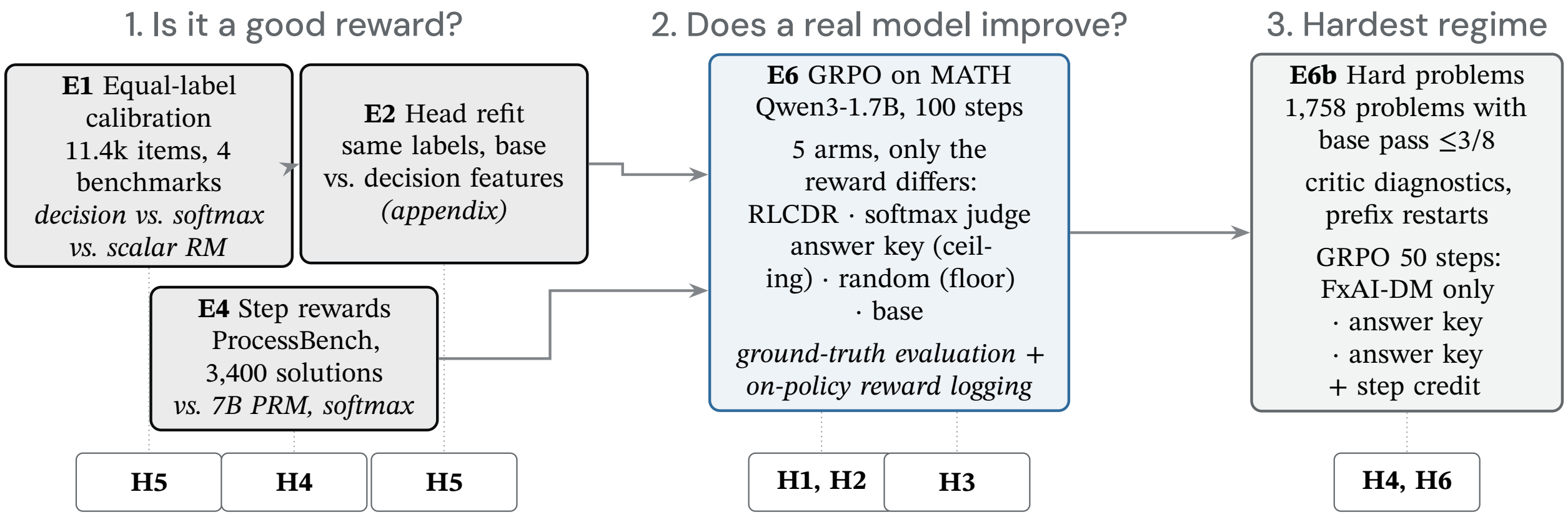

\Cref{fig:roadmap} shows how the experiments fit together.

\section{Related Work}
\label{sec:related}

We cover the four lines of work closest to ours.

\paragraph{LLM judges and generative reward models.} LLM judges \citep{zheng2023mtbench,liu2023geval} and AI feedback \citep{bai2022cai,lee2023rlaif,yuan2024selfreward} made model-based rewards cheap to build. Generative verifiers \citep{zhang2024genrm,ankner2024cloud,liu2025grm,chen2025rmr1,whitehouse2025j1}, rubric rewards \citep{gunjal2025rar,viswanathan2025rlcf,zhou2025ruscarl} and model-based answer checkers \citep{su2025bridge,ma2025generalreasoner,yu2025rlpr} carry RL beyond tasks with an exact checker. Frontier post-training now relies on them. DeepSeek-GRM writes principles and critiques and votes over several samples \citep{liu2025grm}. DeepSeek-V4 drops scalar reward models for a rubric-guided generative reward model that the policy itself serves as \citep{deepseek2026v4}. Kimi K2 uses a self-critique rubric reward based on pairwise comparisons \citep{kimi2025k2}, and Kimi K3 an agentic judge that writes a rubric and ranks candidates in a tournament of binary comparisons \citep{kimi2026k3}. None of these reports whether its judge's scores are calibrated, and the Kimi K2 report names finer handling of calibrated uncertainty as future work \citep{kimi2025k2}. They also generate text, often several samples per answer; our reward is read in one forward pass.

\paragraph{Scalar reward models and overoptimization.} Bradley--Terry reward models \citep{stiennon2020summarize,ouyang2022instructgpt} remain the default, with strong recent instances \citep{wang2024armorm,liu2025skywork}. Overoptimization \citep{gao2022overopt}, length bias \citep{singhal2023length,chen2024odin} and imperfect ensembles \citep{coste2023ensembles,eisenstein2023herding} motivated uncertainty-aware and distributional reward models \citep{zhai2023uprlhf,yang2024bayesrm,lou2024urm,dorka2024qrm,sun2025purm,chen2026oprm}. Reward benchmarks measure accuracy \citep{lambert2024rewardbench,malik2025rewardbench2,liu2024rmbench,frick2024ppe}, but accuracy alone predicts RL outcomes poorly \citep{wen2024rethinking,razin2025teacher}, and RL gains on Qwen models can appear even with random rewards \citep{shao2025spurious}. We therefore evaluate rewards by \emph{training with them}, against a random-reward floor and an answer-key ceiling.

\paragraph{Process reward models.} Process supervision began with step-labelled reward models \citep{uesato2022,lightman2023} and moved to automatic labels from Monte Carlo roll-outs \citep{wang2023shepherd,luo2024omega,setlur2024pav} and implicit process rewards \citep{yuan2024implicit,cui2025prime}. ProcessBench and PRMBench showed that PRMs often miss the first error \citep{zheng2024processbench,song2025prmbench}, and Monte Carlo labels bias them \citep{zhang2025prmlessons}. Generative ``thinking'' verifiers are more accurate at the price of long generations \citep{khalifa2025thinkprm,zhao2025genprm,she2025rprm}. Used as RL rewards, PRMs are hacked \citep{cheng2025pure,deepseek2025r1}; recent fixes co-train or gate the verifier \citep{zha2025tango,zou2025reasonflux,agrawal2026verigate}. Our step reward comes from a general decision model with no step training.

\paragraph{Judge calibration.} LLM judges read through their softmax are overconfident \citep{tian2025overconf}, sensitive to option order \citep{shi2024position,zheng2023pride} and attackable \citep{zhao2025masterkey}. Order effects can be corrected \citep{zhao2021calibrate,holtzman2021surface,zhou2023batchcal}, and confidence can be improved by verbalized estimates \citep{tian2023justask}, linear probes \citep{radharapu2025probes} or distribution matching \citep{li2025autoraters}. Reward miscalibration passes into the trained policy \citep{leng2024taming}. The only calibrated PRMs we know of are calibrated post hoc for one policy and used at inference time \citep{park2025prmcal,ma2026distprm}. Probability readouts have been used as RL rewards before \citep{lee2023rlaif,zhang2024genrm,su2025bridge,shen2026ccr,xie2026probejudge,chen2026oprm,galego2026rljf,li2026visualjev}, but none measures calibration on the policy's own samples during RL (\cref{tab:novelty}). As far as we can determine, the \emph{combination} is new: a debiased, label-calibrated, typed multi-option decision readout used as an online RL reward for both outcomes and steps, evaluated with equal-label, on-policy and random-reward controls.

\begin{table}[t]
\centering\small
\caption{\textbf{Probability readouts used as RL rewards.} \cmark{} = present, \xmark{} = absent. Compiled from the cited papers.}
\label{tab:novelty}
\begin{tabular}{@{}lccccc@{}}
\toprule
Work & Calibrated & Debiased & Typed / multi-option & Step-level & Online LLM RL \\
\midrule
d-RLAIF \citep{lee2023rlaif} & \xmark & labels only & score scale & \xmark & \cmark \\
GenRM \citep{zhang2024genrm} & \xmark & \xmark & binary & \xmark & \xmark \\
Reward Bridge \citep{su2025bridge} & \xmark & \xmark & binary & \xmark & \cmark \\
CCR \citep{shen2026ccr} & \xmark & \xmark & binary & \xmark & \cmark \\
Probe judges \citep{xie2026probejudge} & not measured & n/a & binary & \xmark & \cmark \\
OPRM \citep{chen2026oprm} & \cmark & \xmark & score scale & \xmark & \xmark \\
Calibrated PRM \citep{park2025prmcal} & \cmark{} (post hoc) & \xmark & binary & \cmark & \xmark \\
RLJF \citep{galego2026rljf} & \xmark & \xmark & partly & \xmark & \cmark \\
\midrule
\textbf{\method{} (ours)} & \cmark{} \textbf{(on-policy)} & \cmark & \cmark & \cmark & \cmark \\
\bottomrule
\end{tabular}
\end{table}

\section{Method}
\label{sec:method}

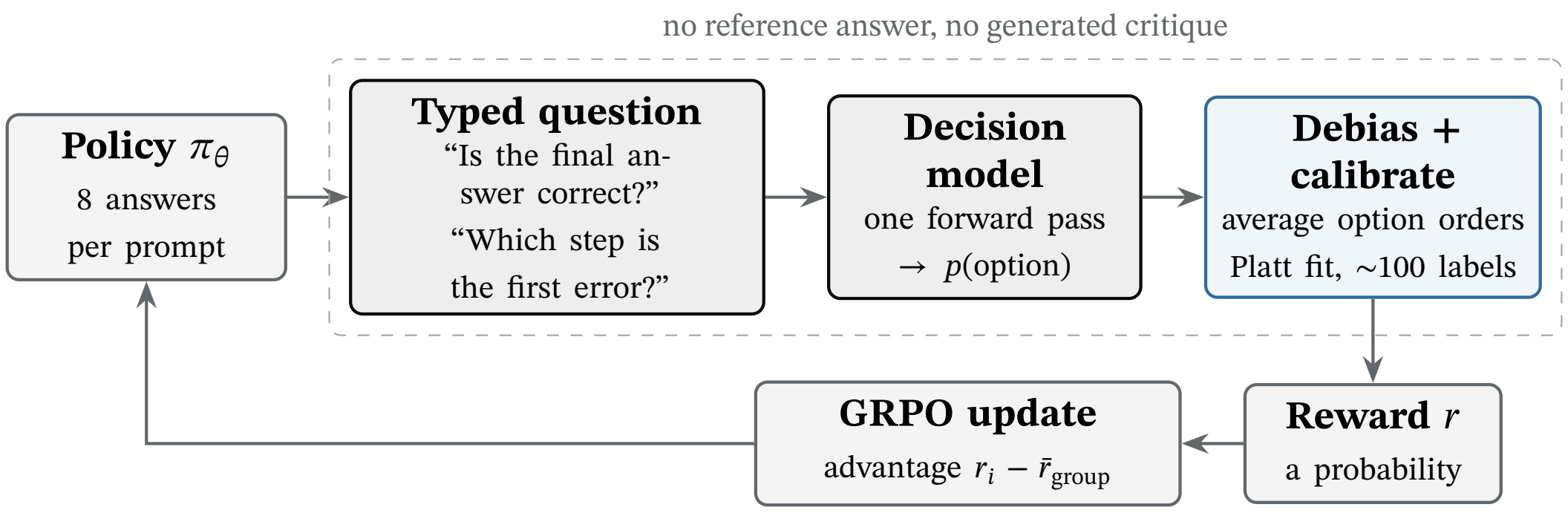
\begin{figure}[t]
\centering
\begin{tikzpicture}[font=\small,
  box/.style={draw=#1,fill=#1!7,rounded corners=3pt,thick,align=center,minimum height=1.05cm,inner sep=4pt},
  arr/.style={-{Stealth[length=2.4mm]},thick,draw=fxgray}]
\node[box=fxgray,text width=2.2cm] (pol) {\textbf{Policy} $\pi_\theta$\\[1pt]\scriptsize 8 answers per prompt};
\node[box=fxblue,text width=3.4cm,right=0.55cm of pol] (q) {\textbf{Typed question}\\[1pt]\scriptsize ``Is the final answer correct?''\\ \scriptsize ``Which step is the first error?''};
\node[box=fxblue,text width=2.5cm,right=0.55cm of q] (dm) {\textbf{Decision model}\\[1pt]\scriptsize one forward pass\\ \scriptsize $\to p(\text{option})$};
\node[box=fxteal,text width=2.7cm,right=0.55cm of dm] (cal) {\textbf{Debias + calibrate}\\[1pt]\scriptsize average option orders\\ \scriptsize Platt fit, $\sim$100 labels};
\node[box=fxorange,text width=2.0cm,below=0.75cm of cal] (r) {\textbf{Reward} $r$\\[1pt]\scriptsize a probability};
\node[box=fxgray,text width=3.5cm,left=0.55cm of r] (grpo) {\textbf{GRPO update}\\[1pt]\scriptsize advantage $r_i - \bar r_{\text{group}}$};
\draw[arr] (pol) -- (q); \draw[arr] (q) -- (dm); \draw[arr] (dm) -- (cal); \draw[arr] (cal) -- (r);
\draw[arr] (r) -- (grpo); \draw[arr] (grpo.west) -| (pol.south);
\node[draw=fxgray!60,dashed,rounded corners=3pt,fit=(q)(dm)(cal),inner sep=5pt,label={[font=\scriptsize\color{fxgray}]above:no reference answer, no generated critique}] {};
\end{tikzpicture}
\caption{\textbf{\method{} in one picture.} Each policy answer is wrapped in a typed question. The decision model returns a distribution over the options in one forward pass. Averaging over option orders removes position bias, and a two-parameter fit on a small labeled set makes the probability calibrated. The calibrated probability is the reward.}
\label{fig:method}
\end{figure}

\paragraph{Decision model.} The decision model (\fx, built by FuturixAI) is a 26.1B-parameter model on a Qwen3.8-27B backbone with a trained readout head. Given a \emph{state} (e.g.\ a problem and a response), an \emph{instruction} and up to 255 named options, the head reads the last hidden state and outputs a categorical distribution $p(\cdot\mid x)$ over the options. No text is generated. The readout was trained for general decision making, not on reward-model data.

\paragraph{Typed questions.} We use two questions, given verbatim in \cref{app:prompts}. The \emph{outcome} question asks whether the response's final answer is correct, with options \{correct, incorrect\}. It gives $p_{\text{ok}}(x)$. The \emph{critic} question presents the response split into steps and asks which step contains the first error, with options \{none, step 1, \dots, step $n$\}. It gives a distribution $q$ over the location of the first error.

\paragraph{Debiasing.} Readouts of multiple-choice questions depend on option order \citep{zheng2023pride,pezeshkpour2023order}. We ask every question in each option order (forward and reversed for the questions here) and average the probability assigned to each option.

\paragraph{Calibration.} We map the debiased probability through a Platt fit, $r = \sigma\!\big(a\cdot\operatorname{logit}(p_{\text{ok}}) + b\big)$. The two parameters are fitted once, by maximum likelihood, on 1,024 labelled policy responses to MATH problems from PPE \citep{frick2024ppe}, collected from \emph{other} models. They are then frozen during RL. The resulting reward is \emph{reference-free}: it never sees the answer key.

\paragraph{Outcome reward and step reward.} The outcome reward is $r_{\text{out}} = r$. From the critic distribution $q$, the \emph{progress} of a response with $n$ steps is the expected fraction of steps that come before the first error:
\begin{equation}
  \operatorname{prog}(x) \;=\; q(\text{none}) \;+\; \sum_{k=1}^{n} q(\text{step }k)\,\frac{k-1}{n}.
\end{equation}
The hard-problem experiments (\cref{sec:hard}) combine the two. The \emph{reference-free} step reward replaces the answer key by the decision model's own outcome probability,
\begin{equation}
  r_{\text{step}}(x) \;=\; r_{\text{out}}(x) + \big(1-r_{\text{out}}(x)\big)\,\lambda\,\operatorname{prog}(x), \qquad \lambda=0.5,
\end{equation}
so an answer judged correct scores near 1 and an answer judged wrong is graded by how far it got. The \emph{hybrid} reward gives 1 to a correct final answer according to the answer key and $\lambda\cdot\operatorname{prog}(x)$ otherwise; it tests step credit alone. A 0/1 checker gives exactly zero gradient whenever all answers in a group fail.

\paragraph{Policy optimization.} We use GRPO \citep{shao2024deepseekmath} with the Dr.~GRPO loss \citep{liu2025drgrpo}: group-mean baseline, no division by the group standard deviation and no length normalization. The advantage of each sample is therefore $r_i - \bar r$, so a calibrated reward enters the gradient on its own scale.

\section{Experimental Setup}
\label{sec:setup}

\begin{table}[t]
\centering\small
\caption{\textbf{Every experiment at a glance.} ``Labels'' is the number of labelled examples any method may use for calibration. All splits are by question; nothing is fitted on evaluation items.}
\label{tab:program}
\begin{tabularx}{\linewidth}{@{}l X >{\raggedright\arraybackslash}p{3.4cm} l >{\raggedright\arraybackslash}p{2.2cm}@{}}
\toprule
Exp. & Data (size) & Compared & Labels & Main metric \\
\midrule
E1 & PPE best-of-K (5,120), JudgeBench (620), RM-Bench (3,981), RewardBench 2 (1,723) & \fx, softmax, Skywork-V2-8B & 0--1,600 & Brier, NLL, AUROC \\
E2 & Same as E1, hidden states saved & 6 head variants & 25--1,600 & Brier \\
E4 & ProcessBench (3,400 solutions, 16,548 steps) & \fx, softmax, Qwen2.5-Math-PRM-7B & 0--400 & First-error F1 \\
E6 & MATH train (5,584) $\to$ MATH500, AIME24/25 & 5 rewards (\cref{sec:setup}) & 1,024 (frozen) & pass@1, pass@$k$ \\
E6b & 1,758 hard MATH problems; 4,451 + 800 policy answers & \fx{} only; answer key $\pm$ step credit & 1,024 (frozen) & pass@1, re-solve drop \\
\bottomrule
\end{tabularx}
\end{table}

\paragraph{Offline protocol (E1, E2, E4).} We use 20 random half splits by question. From the fit half we draw $N$ labelled items, fit each method's calibration on them, and evaluate on the other half. Every method gets the same $N$ items. We report Brier and log-loss (NLL), which are \emph{proper} scoring rules: they reward a probability that is both calibrated and informative. Expected calibration error (ECE, 10 equal-width bins) is reported but never decides, because after calibration even a constant base-rate predictor has ECE 0. AUROC measures ranking alone.

\paragraph{Policy and data.} The policy is Qwen3-1.7B \citep{yang2025qwen3} in non-thinking mode. It trains on the MATH training split (levels 3--5, 5,584 problems), with the prompt ``Please reason step by step, and put your final answer within \texttt{\textbackslash boxed\{\}}.'' We removed every training problem whose normalized text matches an evaluation problem.

\paragraph{RL recipe (identical for every run).} The recipe uses TRL's GRPO trainer with vLLM generation:
\begin{itemize}
  \item 32 prompts $\times$ 8 samples per step; temperature 1.0; up to 3,072 completion tokens.
  \item Learning rate $2\!\times\!10^{-6}$ (constant); KL coefficient $\beta=10^{-3}$; 100 steps.
  \item fp32 master weights with bf16 compute.
\end{itemize}
Each run used one NVIDIA RTX PRO 6000 (96\,GB) for the policy, plus a second one for the reward model when needed. \Cref{app:hparams} lists every setting.

\paragraph{Rewards compared.} We compare five rewards, and only the reward differs between runs:
\begin{itemize}
  \item \textbf{\method{}}: the calibrated \fx{} outcome reward.
  \item \textbf{Softmax judge}: the \emph{same} Qwen3.8-27B backbone, the \emph{byte-identical} prompt, the same debiasing and the same calibration procedure, but reading the next-token softmax over the option letters (the G-Eval / d-RLAIF readout \citep{liu2023geval,lee2023rlaif}). It has the same size and knowledge as our reward and differs only in the readout, which isolates the contribution of decision training.
  \item \textbf{Answer key}: a math-verify check against the reference answer. This is the ceiling.
  \item \textbf{Random}: Bernoulli(0.5). This is the floor, and it catches gains that come from RL itself rather than from the reward \citep{shao2025spurious}.
  \item \textbf{Base}: no training.
\end{itemize}
The answer-key check is computed and logged for every run, but has zero weight except in the answer-key run. This lets us measure the reward's faithfulness on-policy.

\paragraph{Evaluation.} Evaluation uses reference answers only, never a reward model:
\begin{itemize}
  \item MATH500 with 4 samples per problem; AIME 2024 and AIME 2025 with 16 samples each.
  \item Temperature 0.6, top-$p$ 0.95.
  \item We report pass@1 (mean accuracy) and pass@$k$. ``MATH500-clean'' drops the 8 problems that also appear in the calibration set.
\end{itemize}
Differences are \emph{paired by problem}. We report 95\% bootstrap intervals (10,000 resamples) and paired $t$-test $p$-values. Evaluation itself is noisy: re-evaluating the same checkpoint gave 74.85 and 73.90 on MATH500, because vLLM batching is not deterministic. We therefore treat differences of 1--2 points as unresolved unless their paired interval excludes zero, and we do not claim single significant results on the 30-problem AIME sets, where we run many comparisons.

\section{Results}
\label{sec:results}

\subsection{H1--H2: The policy improves, and the reward is the cause}
\label{sec:rl}

\begin{table}[t]
\centering\small
\caption{\textbf{Ground-truth evaluation after 100 GRPO steps} (pass@1, \%; MATH500 pass@4 in the last column). Only the reward differs between rows. \best{Bold}: best reward that does not use reference answers.}
\label{tab:main}
\begin{tabular}{@{}lccccc@{}}
\toprule
Reward used for RL & MATH500 & M500-clean & AIME24 & AIME25 & M500 pass@4 \\
\midrule
None (base model) & 72.2 & 72.3 & 11.3 & 9.4 & 85.2 \\
Random (floor) & 72.8 & 72.8 & 12.3 & 8.5 & 85.8 \\
Softmax judge (same backbone) & 75.5 & 75.5 & 12.9 & 10.0 & 88.4 \\
\rowcolor{fxlight}\textbf{\method{} (ours)} & \best{76.8} & \best{76.9} & \best{16.5} & \best{10.8} & \best{88.4} \\
\midrule
\textcolor{fxgray}{Answer key (uses reference answers)} & \textcolor{fxgray}{76.9} & \textcolor{fxgray}{77.1} & \textcolor{fxgray}{14.4} & \textcolor{fxgray}{13.1} & \textcolor{fxgray}{89.2} \\
\bottomrule
\end{tabular}
\end{table}

\begin{table}[t]
\centering\small
\caption{\textbf{Paired differences on MATH500 after 100 steps} (points; 500 problems $\times$ 4 samples). The second block isolates the causes.}
\label{tab:paired}
\begin{tabular}{@{}llrcr@{}}
\toprule
& Comparison & $\Delta$ & 95\% CI & $p$ \\
\midrule
\multirow{4}{*}{\shortstack[l]{Gain over\\base model}}
 & \method{} $-$ base & \up{+4.65} & $[+2.80,\,+6.50]$ & $<10^{-4}$ \\
 & Answer key $-$ base & +4.75 & $[+3.00,\,+6.55]$ & $<10^{-4}$ \\
 & Softmax $-$ base & +3.35 & $[+1.55,\,+5.15]$ & $3\!\times\!10^{-4}$ \\
 & Random $-$ base & +0.65 & $[-1.30,\,+2.50]$ & 0.50 \\
\midrule
\multirow{3}{*}{\shortstack[l]{What causes\\the gain}}
 & \method{} $-$ random & \up{+4.00} & $[+2.05,\,+6.05]$ & $10^{-4}$ \\
 & \method{} $-$ answer key & $-0.10$ & $[-1.75,\,+1.50]$ & 0.90 \\
 & \method{} $-$ softmax & +1.30 & $[-0.50,\,+3.10]$ & 0.16 \\
\bottomrule
\end{tabular}
\end{table}

\Cref{fig:teaser} and \cref{tab:main,tab:paired} answer the question that matters most: \emph{does a real model get better?} It does. Three findings attribute the gain to the reward.

\textbf{(1) The gain is real and large.} \method{} improves MATH500 by +4.65 points ($p<10^{-4}$). The gain is the same on the clean subset (+4.62), so it does not come from overlap with the calibration data. Accuracy rises at each checkpoint we evaluated (72.2 $\to$ 73.9 $\to$ 73.9 $\to$ 75.6 $\to$ 76.8; \cref{fig:teaser}, right). pass@4 also rises (85.2 $\to$ 88.4): the model solves more distinct problems, rather than merely sharpening answers it could already find.

\textbf{(2) The reward causes it, not RL itself.} The random-reward run uses the same data, steps and optimizer and gains only +0.65 (n.s.). Our reward beats it by +4.0 points ($p=10^{-4}$). This rules out the explanation that RL alone improves Qwen math models \citep{shao2025spurious}.

\textbf{(3) It matches the answer key without using it.} \method{} and the answer-key run are statistically identical ($-0.10$, 95\% CI $[-1.75, 1.50]$). The decision model never sees the reference answer, so this level of training signal is available on problems that \emph{have no reference answer}, which is where RL most needs it.

\textbf{Against the softmax judge, the policy-level comparison is unresolved.} Both judges improve the policy. Our reward is ahead on MATH500 (+1.30, 95\% CI $[-0.50,+3.10]$, $p=0.16$), which is within evaluation noise for one seed, so we make no claim about which of the two trains a better policy. As \emph{rewards}, the two differ clearly: ours is better calibrated and ranks answers better on the policy's own samples (\cref{sec:faithful}) and with equal labels offline (\cref{sec:offline}). \Cref{sec:faithful} explains why GRPO turns only part of that difference into policy accuracy.

\textbf{Our reward teaches what the answer key teaches.} The two runs that improved most also changed the policy in the same way (\cref{tab:length}). Both the answer-key run and \method{} learned to reason at greater length, raising mean MATH500 response length from 837 to about 1,050 tokens. The softmax and random runs did not. Because the behaviour appears under the ground-truth reward too, it reflects learning, not exploitation of our reward.

\begin{table}[h]
\centering\small
\caption{\textbf{Response length after 100 steps.} Mean MATH500 tokens; share of responses truncated at 3,072 tokens.}
\label{tab:length}
\begin{tabular}{@{}lccccc@{}}
\toprule
& Base & Random & Softmax & \textbf{\method{}} & Answer key \\
\midrule
MATH500 mean tokens & 837 & 863 & 856 & \textbf{1,056} & 1,049 \\
MATH500 truncated & 4.3\% & 5.0\% & 4.0\% & 9.7\% & 9.2\% \\
AIME24 truncated & 37\% & 32\% & 37\% & 62\% & 64\% \\
\bottomrule
\end{tabular}
\end{table}

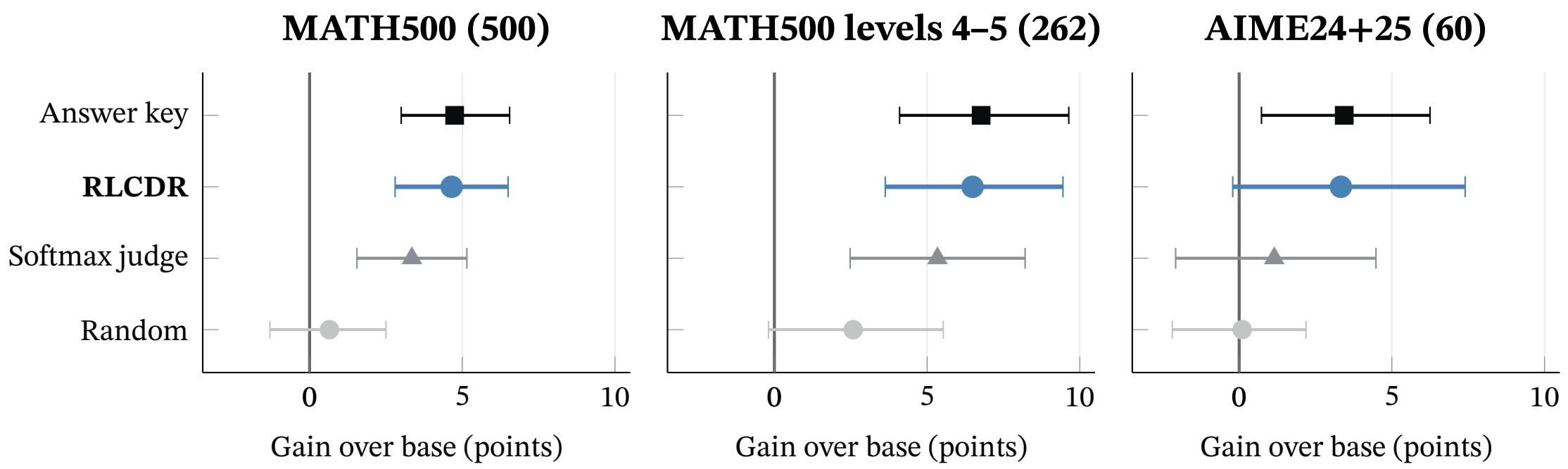
\begin{figure}[t]
\centering
\begin{tikzpicture}
\begin{groupplot}[group style={group size=3 by 1,horizontal sep=0.35cm,yticklabels at=edge left},
  height=4.4cm, width=5.6cm, xmin=-3.5, xmax=10.5, ymin=0.4, ymax=4.6,
  ytick={1,2,3,4}, yticklabels={Random,Softmax judge,\textbf{\method{}},Answer key},
  tick label style={font=\scriptsize}, label style={font=\scriptsize},
  title style={font=\small\bfseries,yshift=-3pt}, xlabel={Gain over base (points)},
  axis x line*=bottom, axis y line*=left, xmajorgrids, grid style={gray!15},
  extra x ticks={0}, extra x tick style={grid=major,grid style={black!60,thick}}]
\nextgroupplot[title={MATH500 (500)}]
\addplot[only marks,mark=*,mark size=2.4pt,cRand,error bars/.cd,x dir=both,x explicit,error bar style={thick}] coordinates {(0.65,1) += (1.85,0) -= (1.95,0)};
\addplot[only marks,mark=triangle*,mark size=2.8pt,cSoft,error bars/.cd,x dir=both,x explicit,error bar style={thick}] coordinates {(3.35,2) += (1.80,0) -= (1.80,0)};
\addplot[only marks,mark=*,mark size=2.8pt,cOurs,error bars/.cd,x dir=both,x explicit,error bar style={very thick}] coordinates {(4.65,3) += (1.85,0) -= (1.85,0)};
\addplot[only marks,mark=square*,mark size=2.3pt,cGT,error bars/.cd,x dir=both,x explicit,error bar style={thick}] coordinates {(4.75,4) += (1.80,0) -= (1.75,0)};
\nextgroupplot[title={MATH500 levels 4--5 (262)}]
\addplot[only marks,mark=*,mark size=2.4pt,cRand,error bars/.cd,x dir=both,x explicit,error bar style={thick}] coordinates {(2.58,1) += (2.95,0) -= (2.77,0)};
\addplot[only marks,mark=triangle*,mark size=2.8pt,cSoft,error bars/.cd,x dir=both,x explicit,error bar style={thick}] coordinates {(5.34,2) += (2.87,0) -= (2.86,0)};
\addplot[only marks,mark=*,mark size=2.8pt,cOurs,error bars/.cd,x dir=both,x explicit,error bar style={very thick}] coordinates {(6.49,3) += (2.96,0) -= (2.86,0)};
\addplot[only marks,mark=square*,mark size=2.3pt,cGT,error bars/.cd,x dir=both,x explicit,error bar style={thick}] coordinates {(6.77,4) += (2.87,0) -= (2.67,0)};
\nextgroupplot[title={AIME24+25 (60)}]
\addplot[only marks,mark=*,mark size=2.4pt,cRand,error bars/.cd,x dir=both,x explicit,error bar style={thick}] coordinates {(0.10,1) += (2.09,0) -= (2.29,0)};
\addplot[only marks,mark=triangle*,mark size=2.8pt,cSoft,error bars/.cd,x dir=both,x explicit,error bar style={thick}] coordinates {(1.15,2) += (3.33,0) -= (3.23,0)};
\addplot[only marks,mark=*,mark size=2.8pt,cOurs,error bars/.cd,x dir=both,x explicit,error bar style={very thick}] coordinates {(3.33,3) += (4.07,0) -= (3.54,0)};
\addplot[only marks,mark=square*,mark size=2.3pt,cGT,error bars/.cd,x dir=both,x explicit,error bar style={thick}] coordinates {(3.44,4) += (2.81,0) -= (2.71,0)};
\end{groupplot}
\end{tikzpicture}
\caption{\textbf{Every gain over the base model, with 95\% paired-bootstrap intervals} (after 100 steps; number of problems in brackets). An interval that does not cross the black zero line is a significant gain. Our reward and the answer key are indistinguishable on every slice; random reward is indistinguishable from no training on every slice.}
\label{fig:forest}
\end{figure}

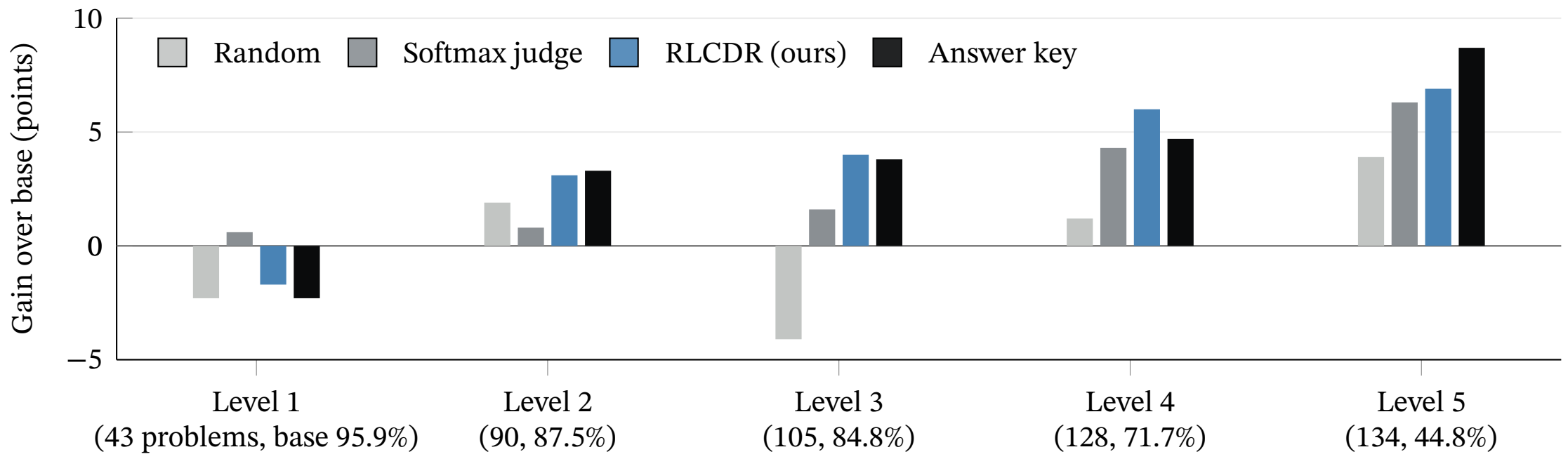
\begin{figure}[t]
\centering
\begin{tikzpicture}
\begin{axis}[ybar, bar width=7pt, height=4.8cm, width=0.92\linewidth, ymin=-5, ymax=10,
  symbolic x coords={L1,L2,L3,L4,L5}, xtick=data,
  xticklabels={{Level 1\\(43 problems, base 95.9\%)},{Level 2\\(90, 87.5\%)},{Level 3\\(105, 84.8\%)},{Level 4\\(128, 71.7\%)},{Level 5\\(134, 44.8\%)}},
  x tick label style={font=\scriptsize,align=center}, tick label style={font=\scriptsize},
  ylabel={\scriptsize Gain over base (points)}, axis x line*=bottom, axis y line*=left,
  ymajorgrids, grid style={gray!15}, enlarge x limits=0.12,
  legend style={font=\scriptsize,at={(0.02,0.98)},anchor=north west,draw=none,legend columns=4,column sep=5pt,fill=white,fill opacity=0.9,text opacity=1},
  extra y ticks={0}, extra y tick style={grid=major,grid style={black!60}}]
\addplot[draw=none,fill=cRand] coordinates {(L1,-2.3)(L2,1.9)(L3,-4.1)(L4,1.2)(L5,3.9)};
\addlegendentry{Random}
\addplot[draw=none,fill=cSoft] coordinates {(L1,0.6)(L2,0.8)(L3,1.6)(L4,4.3)(L5,6.3)};
\addlegendentry{Softmax judge}
\addplot[draw=none,fill=cOurs] coordinates {(L1,-1.7)(L2,3.1)(L3,4.0)(L4,6.0)(L5,6.9)};
\addlegendentry{\method{} (ours)}
\addplot[draw=none,fill=cGT] coordinates {(L1,-2.3)(L2,3.3)(L3,3.8)(L4,4.7)(L5,8.7)};
\addlegendentry{Answer key}
\end{axis}
\end{tikzpicture}
\caption{\textbf{MATH500 gain over base by difficulty level} (after 100 steps). Our reward's gain rises with difficulty, from about +3 points on levels 2--3 to +6.9 on level 5 (44.8\% $\to$ 51.7\%), tracking the answer key. Level 1 has only 43 problems that the base model already solves 96\% of the time, so its differences ($\pm2$ points) are noise.}
\label{fig:levels}
\end{figure}

\paragraph{Where the gain comes from.} \Cref{fig:forest} shows every paired gain over the base model with its 95\% interval, and \cref{fig:levels} splits MATH500 by the dataset's difficulty level. Three things stand out.
\begin{itemize}
  \item \textbf{The gain grows with difficulty.} Our reward adds +2.6 points on levels 1--3 ($p=0.02$), +6.5 on levels 4--5 ($p<10^{-4}$) and +6.9 on level 5 alone ($p=0.002$). The answer-key run shows the same pattern. Gains on easy problems are capped, since the base model already solves 88\% of them.
  \item \textbf{Random reward does not produce this pattern.} Its gains are inside the noise on every slice (levels 4--5: +2.6, $p=0.08$; levels 1--3: $-1.5$).
  \item \textbf{AIME moves in the same direction but is noisier.} Pooled over the 60 AIME problems, our reward gains +3.3 points ($p=0.09$) and the answer key +3.4 ($p=0.02$); the difference between them is $-0.1$ ($p=0.94$). The softmax judge gains +1.2 ($p=0.50$), random +0.1.
\end{itemize}
The model also solves new problems rather than only sharpening old ones: with our reward, 31 MATH500 problems that the base model never solved in 4 attempts become solved, and 15 are lost (answer key: 32 and 12; random: 25 and 22).

\begin{figure}[t]
\centering
\begin{tikzpicture}
\begin{axis}[height=4.6cm, width=0.75\linewidth, xmin=0, xmax=100, ymin=0.6, ymax=0.76,
  xlabel={\scriptsize RL step (10-step windows)}, ylabel={\scriptsize Training-sample accuracy},
  tick label style={font=\scriptsize}, axis x line*=bottom, axis y line*=left, ymajorgrids, grid style={gray!15},
  yticklabel style={/pgf/number format/fixed,/pgf/number format/precision=2},
  legend style={font=\scriptsize,at={(1.03,0.5)},anchor=west,draw=none}]
\addplot[cOurs,very thick,mark=*,mark size=1.6pt] coordinates {(5,0.664)(15,0.688)(25,0.651)(35,0.693)(45,0.668)(55,0.687)(65,0.652)(75,0.723)(85,0.719)(95,0.699)};
\addlegendentry{\method{} (ours)}
\addplot[cGT,thick,dashed,mark=square*,mark size=1.4pt] coordinates {(5,0.660)(15,0.693)(25,0.649)(35,0.680)(45,0.681)(55,0.691)(65,0.670)(75,0.737)(85,0.707)(95,0.696)};
\addlegendentry{Answer key}
\addplot[cSoft,thick,mark=triangle*,mark size=1.8pt] coordinates {(5,0.660)(15,0.693)(25,0.646)(35,0.686)(45,0.684)(55,0.704)(65,0.671)(75,0.739)(85,0.731)(95,0.683)};
\addlegendentry{Softmax judge}
\addplot[cRand,thick,mark=diamond*,mark size=1.8pt] coordinates {(5,0.669)(15,0.685)(25,0.624)(35,0.663)(45,0.636)(55,0.642)(65,0.635)(75,0.683)(85,0.672)(95,0.645)};
\addlegendentry{Random}
\end{axis}
\end{tikzpicture}
\caption{\textbf{Accuracy of the policy's own training samples during RL} (256 samples per step, averaged over 10 steps, labelled by the answer key but never shown to the non-answer-key rewards). Every informative reward pulls away from the random-reward run after about 20 steps. Note: training accuracy is measured on training problems; \cref{tab:main} gives held-out results.}
\label{fig:train}
\end{figure}
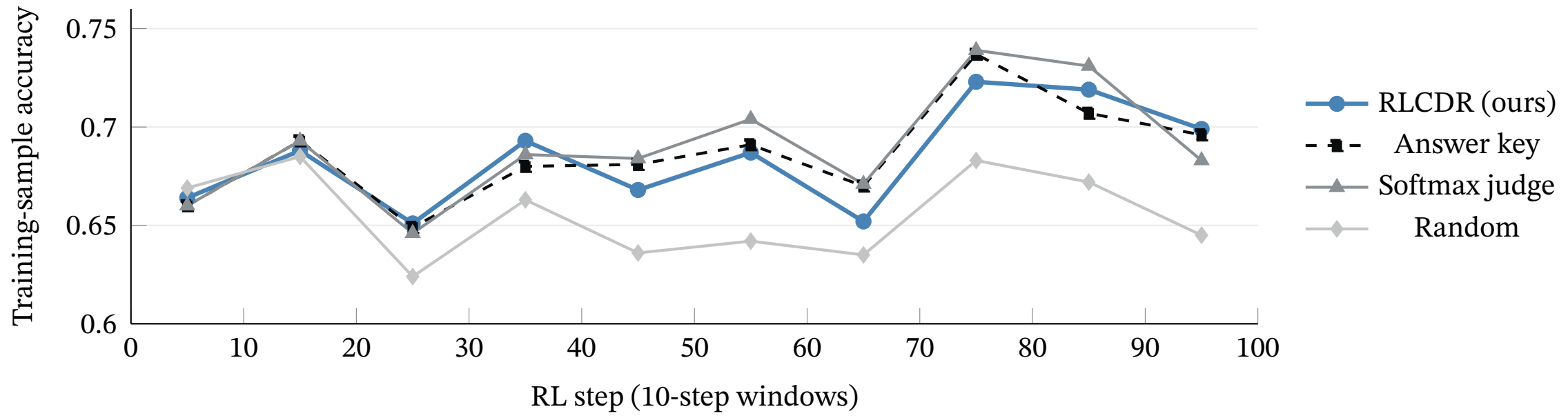

\paragraph{Training dynamics.} Each run logs the answer-key correctness of every training sample (with zero weight, except in the answer-key run). \Cref{fig:train} shows this training accuracy over the 100 steps. All runs see the same prompts in the same order, so the batch-to-batch ups and downs are shared. The three informative rewards climb together, while the random-reward run falls 2--5 points below them from step 20 onward. The policy is learning from the reward signal throughout training, not only at the end.

\subsection{H3: The reward stays honest while the policy changes}
\label{sec:faithful}

A reward that is accurate on a fixed benchmark can still fail during RL: the policy drifts to new kinds of answers and may find ones the reward over-scores. Because every run logs the true correctness of every sample, we can measure the reward's quality \emph{on-policy}, over all 25,600 samples per run.

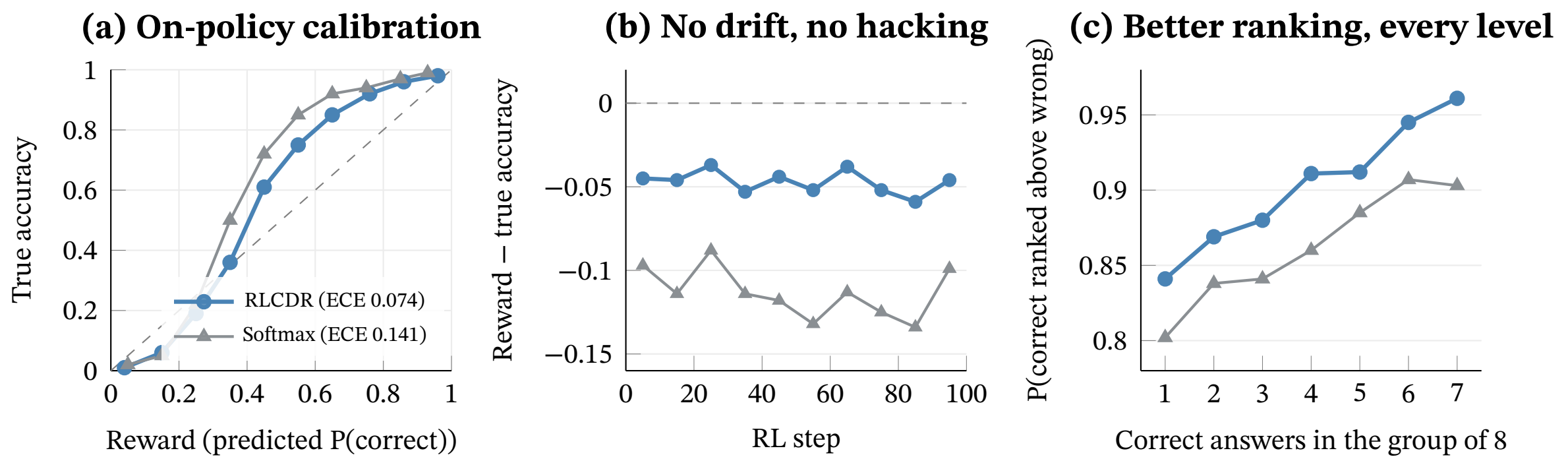
\begin{figure}[t]
\centering
\begin{tikzpicture}
\begin{groupplot}[group style={group size=3 by 1,horizontal sep=1.75cm},
  height=4.6cm, width=5.0cm, tick label style={font=\scriptsize}, label style={font=\scriptsize},
  title style={font=\small\bfseries,yshift=-3pt}, axis x line*=bottom, axis y line*=left, grid style={gray!15}]
\nextgroupplot[xmin=0,xmax=1,ymin=0,ymax=1,xlabel={Reward (predicted P(correct))},ylabel={True accuracy},
  title={(a) On-policy calibration},grid=major,
  legend style={font=\tiny,at={(0.98,0.03)},anchor=south east,draw=none,fill=white,fill opacity=0.85,text opacity=1}]
\addplot[gray,dashed,forget plot] coordinates {(0,0)(1,1)};
\addplot[cOurs,very thick,mark=*,mark size=1.6pt] coordinates {(0.04,0.01)(0.15,0.06)(0.25,0.19)(0.35,0.36)(0.45,0.61)(0.55,0.75)(0.65,0.85)(0.76,0.92)(0.86,0.96)(0.96,0.98)};
\addlegendentry{\method{} (ECE 0.074)}
\addplot[cSoft,thick,mark=triangle*,mark size=1.8pt] coordinates {(0.05,0.02)(0.15,0.05)(0.25,0.22)(0.35,0.50)(0.45,0.72)(0.55,0.85)(0.65,0.92)(0.75,0.94)(0.85,0.97)(0.93,0.99)};
\addlegendentry{Softmax (ECE 0.141)}
\nextgroupplot[xmin=0,xmax=100,ymin=-0.16,ymax=0.02,xlabel={RL step},ylabel={Reward $-$ true accuracy},
  title={(b) No drift, no hacking},ymajorgrids,ytick={-0.15,-0.10,-0.05,0},
  yticklabel style={/pgf/number format/fixed,/pgf/number format/precision=2}]
\addplot[gray,dashed,forget plot] coordinates {(0,0)(100,0)};
\addplot[cOurs,very thick,mark=*,mark size=1.4pt] coordinates {(5,-0.045)(15,-0.046)(25,-0.037)(35,-0.053)(45,-0.044)(55,-0.052)(65,-0.038)(75,-0.052)(85,-0.059)(95,-0.046)};
\addplot[cSoft,thick,mark=triangle*,mark size=1.6pt] coordinates {(5,-0.097)(15,-0.114)(25,-0.088)(35,-0.114)(45,-0.118)(55,-0.132)(65,-0.113)(75,-0.125)(85,-0.134)(95,-0.099)};
\nextgroupplot[xmin=0.5,xmax=7.5,ymin=0.78,ymax=0.98,xtick={1,...,7},xlabel={Correct answers in the group of 8},
  ylabel={P(correct ranked above wrong)},title={(c) Better ranking, every level},ymajorgrids,
  yticklabel style={/pgf/number format/fixed,/pgf/number format/precision=2}]
\addplot[cOurs,very thick,mark=*,mark size=1.6pt] coordinates {(1,0.841)(2,0.869)(3,0.880)(4,0.911)(5,0.912)(6,0.945)(7,0.961)};
\addplot[cSoft,thick,mark=triangle*,mark size=1.8pt] coordinates {(1,0.802)(2,0.838)(3,0.841)(4,0.860)(5,0.885)(6,0.907)(7,0.903)};
\end{groupplot}
\end{tikzpicture}
\caption{\textbf{The reward measured on the policy's own samples during RL} (25,600 samples per run, labelled with the answer key, which neither reward saw). \textbf{(a)} Our reward's probabilities track the true accuracy of the policy's answers about twice as closely. \textbf{(b)} The gap between reward and truth stays flat over training. Our reward is slightly cautious (never overpaying), while the softmax judge undervalues correct answers about twice as much. \textbf{(c)} Within each GRPO group, our reward ranks a correct answer above a wrong one more often at \emph{every} difficulty level. This within-group ordering is what drives GRPO's update.}
\label{fig:faithful}
\end{figure}

\paragraph{Calibration survives policy shift.} On-policy, \method{} has ECE 0.074 and Brier 0.077, against 0.141 and 0.106 for the softmax judge on the same backbone (\cref{fig:faithful}a). It also separates correct from incorrect answers better (AUROC 0.956 vs.\ 0.943). This is not circular. Both rewards were calibrated once, on PPE answers written by four \emph{other} models (Gemma-2-9B, Llama-3-8B, GPT-4o-mini, Claude-3-Haiku), and then frozen. The on-policy numbers are measured on Qwen3-1.7B's own, changing answers, which neither reward saw during calibration. The decision readout carried its calibration onto that new policy.

\paragraph{No reward hacking.} The reward--truth gap of \method{} stays between $-0.04$ and $-0.06$ for all 100 steps (\cref{fig:faithful}b). Its average reward never exceeded the policy's true accuracy, and response length grew without any rise in reward. The softmax judge stays about twice as far from the truth ($-0.09$ to $-0.13$).

\paragraph{Better within-group ranking.} GRPO learns from differences \emph{within} a group of answers to the same prompt. In every group size where some answers are right and some wrong, \method{} ranks right above wrong more often (\cref{fig:faithful}c). This includes the hardest case, groups with only one correct answer. There, \method{} ranks that answer highest 59.3\% of the time, against 52.3\% for the softmax judge (12.5\% by chance), and pushes 47\% more advantage mass toward it.

\paragraph{Why a better reward only partly becomes a better policy.} 57\% of all training groups had either all 8 answers right or all 8 wrong. In those groups both rewards give the policy almost no usable signal. The ranking advantage in \cref{fig:faithful}c therefore applies to only a minority of updates. This points to the regime where calibrated decision rewards should matter most: hard problems.

\subsection{H2, H4, H6: Hard problems, where the answer key is silent}
\label{sec:hard}

On a problem where every sampled answer is wrong, a 0/1 answer checker returns all zeros and GRPO learns nothing. On the 5,584 MATH training problems, the base policy solves 0 of 8 samples on 823 problems and at most 3 of 8 on 1,758. A step-level reward can still tell a nearly-right attempt from a hopeless one.

\paragraph{Does the critic find real errors in our policy's answers?} We asked both 27B judges the critic question on 4,451 policy answers to these hard problems. On these answers the decision readout separates correct from incorrect solutions much better (AUROC of $q(\text{none})$: \textbf{0.781} vs.\ 0.698). It also wrongly declares ``no error'' in a failed solution less often (9.6\% vs.\ 14.4\%).

To check \emph{localization} without human labels, we use roll-outs in the style of Math-Shepherd \citep{wang2023shepherd}. For 800 failed answers, the base policy re-solves the problem four times from just before the step the judge flags, and four times from just after it. If the flagged step really is the error, success should drop once it is included.

\begin{figure}[t]
\centering
\begin{minipage}[c]{0.50\linewidth}
\centering
\begin{tikzpicture}
\begin{axis}[ybar, bar width=20pt, height=4.7cm, width=\linewidth, ymin=-0.025, ymax=0.09,
  symbolic x coords={Random step,Softmax,Ours}, xtick={Random step,Softmax,Ours}, xticklabels={Random step,Softmax judge,\method{} (ours)}, tick label style={font=\scriptsize},
  bar shift=0pt, scaled y ticks=false, error bars/error bar style={black,thick},
  ylabel={\scriptsize Drop in re-solve success}, axis x line*=bottom, axis y line*=left,
  ymajorgrids, grid style={gray!15}, enlarge x limits=0.28,
  yticklabel style={/pgf/number format/fixed,/pgf/number format/precision=2},
]
\addplot+[draw=none,fill=cRand,error bars/.cd,y dir=both,y explicit] coordinates {(Random step,-0.006) +- (0,0.012)};
\addplot+[draw=none,fill=cSoft,error bars/.cd,y dir=both,y explicit] coordinates {(Softmax,0.044) +- (0,0.016)};
\addplot+[draw=none,fill=cOurs,error bars/.cd,y dir=both,y explicit] coordinates {(Ours,0.064) +- (0,0.017)};
\end{axis}
\end{tikzpicture}
\end{minipage}\hfill
\begin{minipage}[c]{0.46\linewidth}
\small
\begin{tabular}{@{}lccc@{}}
\toprule
Flagged step from & Before & After & Drop \\
\midrule
Random step & 0.091 & 0.097 & $-0.006$ \\
Softmax judge & 0.106 & 0.061 & 0.044 \\
\textbf{\method{}} & \textbf{0.133} & 0.069 & \textbf{0.064} \\
\bottomrule
\end{tabular}\\[4pt]
\footnotesize Re-solve success rate from just before vs.\ just after the flagged step (800 failed answers, 4 roll-outs each). Paired on the 668 answers both judges flagged: \method{} $-$ softmax $=+0.016$, 95\% CI $[0.001, 0.031]$.
\end{minipage}
\caption{\textbf{The decision model finds the step where a failed solution goes wrong.} Including the step it flags roughly halves the chance of recovering, whereas a random step has no effect. The effect is 45\% larger than for the softmax judge, and the prefix our model keeps \emph{before} the flagged step is more useful (13.3\% vs.\ 10.6\% recovery).}
\label{fig:prefix}
\end{figure}

\Cref{fig:prefix} shows that it does. Re-solving succeeds about half as often once the step \fx{} flags is included (0.133 $\to$ 0.069), while a random step makes no difference. The effect is also larger than for the softmax judge (paired difference $+0.016$, 95\% CI $[0.001,0.031]$). On these failed attempts the critic's progress score is spread out (mean 0.51, s.d.\ 0.19), so it gives GRPO a usable signal exactly where the answer key gives none.

\paragraph{RL on hard problems.} We trained three policies on the 1,758 hard problems for 50 steps with the recipe of \cref{sec:rl}, changing only the reward (\cref{tab:hardrewards}). The first uses the answer key alone. The second uses the reference-free step reward of \cref{sec:method}, which never sees the answer key. The third, the hybrid, adds \fx{} step credit to the answer key for failed attempts. The second run asks whether the decision model can replace the answer key where the answer key is least informative. The third asks whether step credit adds signal that the answer key lacks.

\begin{table}[h]
\centering\small
\caption{\textbf{What each hard-problem reward sees.} $r_{\text{out}}$ is the calibrated \fx{} probability that the final answer is correct; $\operatorname{prog}$ is the critic's progress score (\cref{sec:method}).}
\label{tab:hardrewards}
\begin{tabularx}{\linewidth}{@{}>{\raggedright\arraybackslash}p{3.0cm} *{3}{>{\centering\arraybackslash}X}@{}}
\toprule
& Answer key & \textbf{\fx{} only (reference-free)} & Answer key + \fx{} step credit \\
\midrule
Score of an answer & 1 if correct, else 0 & $r_{\text{out}} + (1-r_{\text{out}})\,0.5\,\operatorname{prog}$ & 1 if correct, else $0.5\,\operatorname{prog}$ \\
Group with all 8 answers wrong & no gradient & graded & graded \\
Uses the reference answer & yes & \textbf{no} & yes (0/1 part only) \\
\bottomrule
\end{tabularx}
\end{table}

\begin{table}[h]
\centering
\small
\caption{\textbf{RL on hard problems} (pass@1, \%, after 50 steps; paired 95\% bootstrap CIs; level 5 has 134 MATH500 problems). Pooled over both AIME sets, only the hybrid policy improves significantly on base ($+2.7$ $[+1.0,+4.7]$, $p=0.006$; answer key $+1.7$, $p=0.26$; \fx{} only $+1.2$, $p=0.24$).}
\label{tab:hard}
\begin{tabular}{@{}lcccc@{}}
\toprule
Policy (50 steps, hard problems only) & MATH500 & Level 5 & AIME24 & AIME25 \\
\midrule
Base (Qwen3-1.7B) & 72.2 & 44.8 & 11.3 & 9.4 \\
Answer key & 75.3 & 50.6 & \textbf{14.8} & 9.2 \\
\rowcolor{fxlight}\textbf{\fx{} only (reference-free)} & 74.1 & 48.9 & 13.1 & 10.0 \\
Answer key + \fx{} step credit & \textbf{75.7} & \textbf{52.1} & 12.9 & \textbf{13.1} \\
\midrule
\fx{} only $-$ base & $+1.9$ $[+0.2,+3.8]$ & $+4.1$ $[-0.2,+8.4]$ & $+1.9$ & $+0.6$ \\
\fx{} only $-$ answer key & $-1.2$ $[-2.9,+0.5]$ & $-1.7$ $[-5.8,+2.4]$ & $-1.7$ & $+0.8$ \\
Hybrid $-$ answer key & $+0.4$ $[-1.3,+2.1]$ & $+1.5$ $[-2.4,+5.4]$ & $-1.9$ & $+4.0$ $[-0.2,+9.0]$ \\
\bottomrule
\end{tabular}
\end{table}

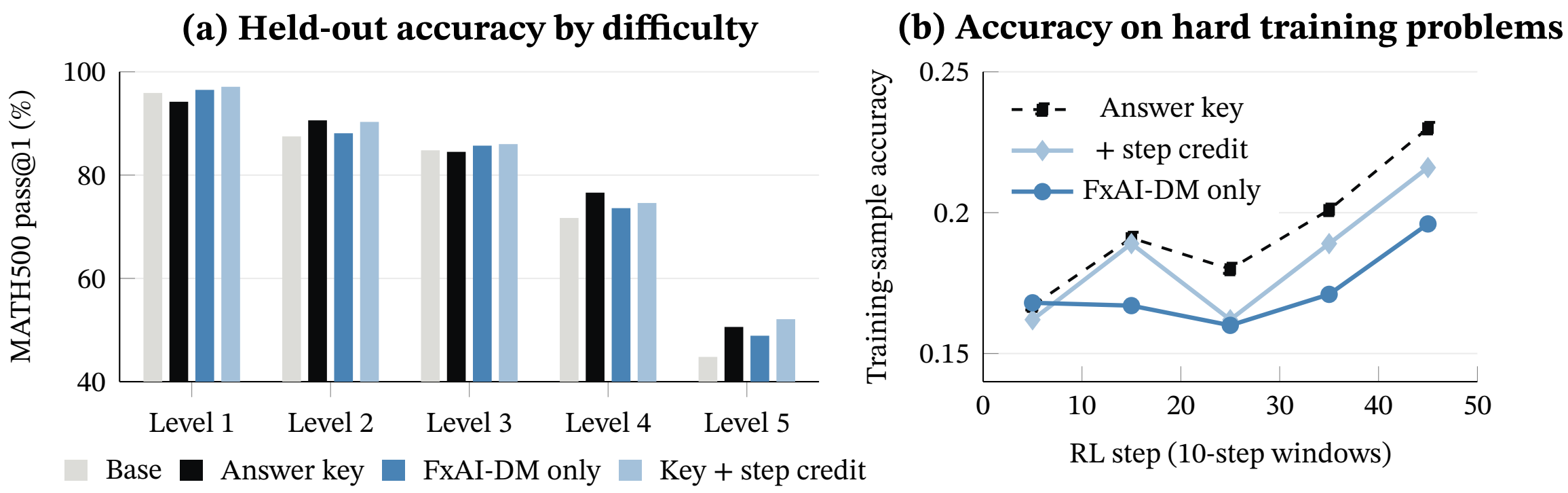
\begin{figure}[t]
\centering
\begin{tikzpicture}
\begin{groupplot}[group style={group size=2 by 1,horizontal sep=1.6cm},
  height=4.6cm, tick label style={font=\scriptsize}, label style={font=\scriptsize},
  title style={font=\small\bfseries,yshift=-3pt}, axis x line*=bottom, axis y line*=left, ymajorgrids, grid style={gray!15}]
\nextgroupplot[width=8.4cm, ybar, bar width=5.2pt, ymin=40, ymax=100,
  symbolic x coords={L1,L2,L3,L4,L5}, xtick=data, xticklabels={Level 1,Level 2,Level 3,Level 4,Level 5},
  ylabel={MATH500 pass@1 (\%)}, title={(a) Held-out accuracy by difficulty}, enlarge x limits=0.13,
  legend style={font=\scriptsize,at={(0.5,-0.2)},anchor=north,draw=none,legend columns=4,column sep=3pt},
  legend image code/.code={\fill[#1] (0cm,-0.08cm) rectangle (0.22cm,0.12cm);}]
\addplot[draw=none,fill=cBase] coordinates {(L1,95.9)(L2,87.5)(L3,84.8)(L4,71.7)(L5,44.8)};
\addlegendentry{Base}
\addplot[draw=none,fill=cGT] coordinates {(L1,94.2)(L2,90.6)(L3,84.5)(L4,76.6)(L5,50.6)};
\addlegendentry{Answer key}
\addplot[draw=none,fill=cOurs] coordinates {(L1,96.5)(L2,88.1)(L3,85.7)(L4,73.6)(L5,48.9)};
\addlegendentry{\fx{} only}
\addplot[draw=none,fill=cHyb] coordinates {(L1,97.1)(L2,90.3)(L3,86.0)(L4,74.6)(L5,52.1)};
\addlegendentry{Key + step credit}
\nextgroupplot[width=6.4cm, xmin=0, xmax=50, ymin=0.14, ymax=0.25, xtick={0,10,20,30,40,50},
  xlabel={RL step (10-step windows)}, ylabel={Training-sample accuracy}, title={(b) Accuracy on hard training problems},
  yticklabel style={/pgf/number format/fixed,/pgf/number format/precision=2},
  legend style={font=\scriptsize,at={(0.03,0.97)},anchor=north west,draw=none}]
\addplot[cGT,thick,dashed,mark=square*,mark size=1.6pt] coordinates {(5,0.167)(15,0.191)(25,0.180)(35,0.201)(45,0.230)};
\addlegendentry{Answer key}
\addplot[cHyb,very thick,mark=diamond*,mark size=2pt] coordinates {(5,0.162)(15,0.189)(25,0.162)(35,0.189)(45,0.216)};
\addlegendentry{+ step credit}
\addplot[cOurs,very thick,mark=*,mark size=1.8pt] coordinates {(5,0.168)(15,0.167)(25,0.160)(35,0.171)(45,0.196)};
\addlegendentry{\fx{} only}
\end{groupplot}
\end{tikzpicture}
\caption{\textbf{Hard-problem RL (50 steps on the 1,758 problems the base model mostly fails).} \textbf{(a)} On held-out MATH500, the reference-free reward (dark blue) is above base at every level but below the answer key on levels 2, 4 and 5. Answer key + step credit (light blue) gives the best level-5 accuracy (52.1\% vs.\ 50.6\%; base 44.8\%). \textbf{(b)} On the training problems, the reference-free run learns more slowly and rises only in the last 20 steps. The hybrid is slightly behind the answer key here, so its held-out advantage is not from fitting the training set better.}
\label{fig:hard}
\end{figure}

\paragraph{Without the answer key, the policy improves but trails.} Trained without any reference answer, the policy still improves: $+1.9$ MATH500 points over base ($[+0.2,+3.8]$, $p=0.03$) and $+4.1$ on level 5 ($p=0.07$). It trails the answer-key policy by 1.2 points on MATH500 ($[-2.9,+0.5]$, $p=0.17$) and by 0.4 on pooled AIME ($p=0.72$). Neither gap is significant with one seed. But the direction is consistent, and accuracy on the training problems agrees (\cref{fig:hard}b: 19.6\% vs.\ 23.0\% over the last 10 steps). We therefore read it as a real shortfall: on hard problems, the reference-free reward is weaker than the answer key, whereas on the full MATH training set it matched it (\cref{sec:rl}).

\paragraph{Why: on hard problems the reward becomes overconfident.} On the policy's own samples in this run, the decision model still separates right from wrong answers well (AUROC 0.89). Its probabilities, however, are too high: the mean reward probability is 0.28 against a true accuracy of 0.17 (ECE 0.11, against 0.074 on the full training set, where it was slightly \emph{under}-confident). At a 0.5 threshold it recognized only 65\% of correct answers. Correct answers therefore received an average reward of 0.73 instead of 1, and wrong answers 0.41. That margin is far smaller than the answer key's 1 against 0. The two Platt parameters were fitted on PPE answers of mixed difficulty, and they do not transfer to a distribution in which 83\% of answers are wrong. Recalibrating on about a hundred labelled hard answers is the direct fix (\cref{sec:limits}).

\paragraph{No hacking, and the gain is in reliability.} The share of wrong answers judged correct stayed at 9--10\% throughout training. The reward on wrong answers was \emph{negatively} correlated with their length ($-0.48$), and answers did not grow (833 MATH500 tokens, base 837). pass@4 was unchanged (85.0\% vs.\ 85.2\%). The gain is in pass@1: the policy became more reliable on problems it could already solve sometimes, rather than solving new ones. At step 25 it was still at base level (72.3\%), so all of the gain came in the second half, and training accuracy was still rising when we stopped.

\paragraph{Step credit works best added to the answer key.} The hybrid reward gives the best hard-problem policy. It reaches 75.7\% on MATH500 and 52.1\% on level 5, $+7.3$ over base ($[+3.0,+11.9]$, $p=0.002$) against $+5.8$ for the answer key ($p=0.01$). It is also the only run with a significant pooled-AIME gain over base. Head-to-head against the answer key, the differences are not significant ($+0.4$ MATH500, $p=0.65$; $+1.5$ level 5, $p=0.46$; $+1.0$ pooled AIME, $p=0.50$), so we claim parity with a positive trend. During training, 42--51\% of the hybrid run's groups had all answers wrong; these received graded credit instead of zero gradient. Partial credit was negatively correlated with length ($-0.4$), and the hybrid's answers were shorter (921 vs.\ 981 tokens; 5.7\% vs.\ 7.8\% truncated). On the training problems themselves the answer-key run is slightly ahead (23.0\% vs.\ 21.6\% over the last 10 steps), so the held-out advantage does not come from fitting the training set better.

\subsection{H4: Step-level rewards, better than a dedicated process reward model}
\label{sec:steps}

\begin{table}[t]
\centering\small
\caption{\textbf{Finding the first wrong step} (ProcessBench F1, 3,400 human-labelled solutions; 20 random half splits; ``labels'' = solutions used to fit the calibration and the cutoff). \fx{} was never trained on step labels.}
\label{tab:processbench}
\begin{tabular}{@{}lcccccc@{}}
\toprule
Method & Labels & GSM8K & MATH & Olympiad & Omni-MATH & \textbf{Average} \\
\midrule
Softmax judge (same 27B backbone) & 100 & 0.725 & 0.721 & 0.644 & 0.637 & 0.682 \\
Qwen2.5-Math-PRM-7B (step-trained) & 100 & 0.811 & 0.773 & 0.659 & 0.645 & 0.722 \\
\rowcolor{fxlight}\textbf{\fx{} decision (ours)} & 100 & \best{0.816} & \best{0.799} & \best{0.736} & \best{0.737} & \best{0.772} \\
\midrule
\fx{} decision, zero-shot & 0 & 0.858 & 0.799 & 0.705 & 0.716 & 0.770 \\
\fx{} one-question critic, zero-shot & 0 & 0.861 & 0.789 & 0.719 & 0.687 & 0.764 \\
Softmax one-question critic, zero-shot & 0 & 0.790 & 0.731 & 0.655 & 0.610 & 0.696 \\
\bottomrule
\end{tabular}
\end{table}

\Cref{tab:processbench} evaluates step rewards on ProcessBench \citep{zheng2024processbench}. Our reproduction of Qwen2.5-Math-PRM-7B \citep{zhang2025prmlessons} scores 73.0 on average (published: 73.5). The decision model asks ``is step $k$ correct?'' over the prefix, as a PRM would during RL. It outperforms the PRM, which was trained for exactly this task, by +5 F1 on average and in \textbf{20/20} splits. The margin is largest on the hardest sets (Olympiad +7.7, Omni-MATH +9.2). Its step probabilities are also better: AUROC 0.972 vs.\ 0.954, and Brier 0.041 vs.\ 0.051 after calibration. Calibration lowers the decision model's step-level ECE from 0.073 to 0.011 with 100 labels. The one-question critic used for our RL reward reaches 0.764 zero-shot in a single pass per solution, against 0.696 for the same question read through softmax.

\begin{figure}[t]
\centering
\begin{minipage}[c]{0.56\linewidth}
\centering
\begin{tikzpicture}
\begin{axis}[ybar, bar width=7pt, height=4.6cm, width=\linewidth, ymin=0.55, ymax=0.88,
  symbolic x coords={GSM8K,MATH,Olympiad,Omni,Avg}, xtick=data,
  xticklabels={GSM8K,MATH,Olympiad,Omni-MATH,\textbf{Average}},
  tick label style={font=\scriptsize}, ylabel={\scriptsize ProcessBench F1},
  axis x line*=bottom, axis y line*=left, ymajorgrids, grid style={gray!15}, enlarge x limits=0.13,
  yticklabel style={/pgf/number format/fixed,/pgf/number format/precision=2},
  legend style={font=\tiny,at={(0.5,1.02)},anchor=south,draw=none,legend columns=3,column sep=3pt}]
\addplot[draw=none,fill=cSoft] coordinates {(GSM8K,0.725)(MATH,0.721)(Olympiad,0.644)(Omni,0.637)(Avg,0.682)};
\addlegendentry{Softmax (27B)}
\addplot[draw=none,fill=cPRM] coordinates {(GSM8K,0.811)(MATH,0.773)(Olympiad,0.659)(Omni,0.645)(Avg,0.722)};
\addlegendentry{Math-PRM-7B}
\addplot[draw=none,fill=cOurs] coordinates {(GSM8K,0.816)(MATH,0.799)(Olympiad,0.736)(Omni,0.737)(Avg,0.772)};
\addlegendentry{\fx{} (ours)}
\end{axis}
\end{tikzpicture}
\end{minipage}\hfill
\begin{minipage}[c]{0.42\linewidth}
\footnotesize\setlength{\tabcolsep}{4pt}
\begin{tabular}{@{}lccc@{}}
\toprule
Step probabilities & AUROC$\uparrow$ & Brier$\downarrow$ & ECE \\
\midrule
Softmax, 0 labels & 0.940 & 0.122 & 0.211 \\
Softmax, 100 & 0.940 & 0.058 & 0.015 \\
Math-PRM-7B, 100 & 0.954 & 0.051 & 0.012 \\
\textbf{\fx{}, 0 labels} & \textbf{0.972} & 0.051 & 0.073 \\
\textbf{\fx{}, 100} & \textbf{0.972} & \textbf{0.041} & \textbf{0.011} \\
\bottomrule
\end{tabular}\\[4pt]
\footnotesize All 16,548 labelled steps. \fx{} beats the PRM's Brier in 20/20 splits at 100 and 400 labels.
\end{minipage}
\caption{\textbf{Step-level rewards on ProcessBench} (100 calibration labels, mean of 20 splits). The decision model, never trained on step labels, beats the dedicated step-trained PRM on every subset, and its lead is largest on the hardest ones (Olympiad +7.7, Omni-MATH +9.2 F1). Its step probabilities also rank better (AUROC) and, after calibration, are the most accurate (Brier).}
\label{fig:pb}
\end{figure}

\subsection{H5: Calibrated decisions vs.\ softmax and scalar rewards, given equal labels}
\label{sec:offline}

\begin{figure}[t]
\centering
\begin{minipage}[c]{0.44\linewidth}
\centering
\begin{tikzpicture}
\begin{axis}[height=4.6cm,width=\linewidth,xmode=log,log basis x=10,xmin=15,xmax=2500,ymin=0.17,ymax=0.25,
  xtick={25,100,400,1600},xticklabels={25,100,400,1600},
  extra x ticks={15},extra x tick labels={0},
  xlabel={\scriptsize Calibration labels},ylabel={\scriptsize Held-out Brier $\downarrow$},
  tick label style={font=\scriptsize},axis x line*=bottom,axis y line*=left,ymajorgrids,grid style={gray!15},
  yticklabel style={/pgf/number format/fixed,/pgf/number format/precision=2},
  legend style={font=\tiny,at={(0.5,-0.36)},anchor=north,draw=none,legend columns=3,column sep=3pt}]
\addplot[cOurs,very thick,mark=*,mark size=1.6pt] coordinates {(15,0.188)(25,0.194)(100,0.184)(400,0.179)(1600,0.179)};
\addlegendentry{\fx{} decision}
\addplot[cSoft,thick,mark=triangle*,mark size=1.8pt] coordinates {(15,0.227)(25,0.239)(100,0.222)(400,0.217)(1600,0.217)};
\addlegendentry{Softmax}
\addplot[cScalar,thick,mark=square*,mark size=1.4pt] coordinates {(25,0.224)(100,0.217)(400,0.212)(1600,0.210)};
\addlegendentry{Scalar RM}
\end{axis}
\end{tikzpicture}
\end{minipage}\hfill
\begin{minipage}[c]{0.53\linewidth}
\small
\begin{tabular}{@{}lccc@{}}
\toprule
Brier $\downarrow$ ($N{=}100$) & \fx{} & Softmax & Scalar RM \\
\midrule
PPE (policy answers) & \best{0.184} & 0.222 & 0.217 \\
JudgeBench & \best{0.297} & 0.373 & 0.352 \\
RM-Bench & 0.160 & 0.219 & \best{0.079} \\
RewardBench 2 & 0.214 & 0.215 & \best{0.153} \\
\bottomrule
\end{tabular}\\[4pt]
\footnotesize 20 random splits; every method gets the same $N$ labels. \fx{} beats the same-backbone softmax judge in 20/20 splits on PPE, JudgeBench and RM-Bench, and ties on RewardBench 2. Scalar RM: Skywork-Reward-V2-Llama-3.1-8B \citep{liu2025skywork}, trained on 26M preference pairs.
\end{minipage}
\caption{\textbf{Equal-label comparison of reward probabilities.} \textbf{Left:} on PPE (one sampled policy answer per item, the RL reward case), the decision readout with \emph{zero} labels is already better than either baseline with 1,600 labels. \textbf{Right:} the decision readout wins where the question is whether an answer is correct. A preference-trained scalar RM wins on preference-pair benchmarks that match its training format.}
\label{fig:offline}
\end{figure}

\Cref{fig:offline} compares reward probabilities on four benchmarks \citep{frick2024ppe,tan2024judgebench,liu2024rmbench,malik2025rewardbench2}, giving every method the same calibration labels and judging with proper scoring rules. After calibration any method can reach low ECE (even a constant base-rate predictor has ECE 0), so ECE alone cannot decide. Brier and log-loss reward a probability that is both calibrated and informative.

The decision readout is the better reward probability for \emph{judging correctness}. On PPE it reaches AUROC 0.809 vs.\ 0.709 for softmax and 0.708 for the scalar RM, with the largest margins on MATH (0.877) and IFEval (0.829). It is also about 5$\times$ less position-biased than the softmax readout. Without debiasing, the softmax judge loses 5.4 accuracy points on PPE and 8.8 on RM-Bench, while the decision readout loses at most 0.8. The scalar RM's raw $\sigma(r)$ is far from a probability: it predicts 0.90 where the true rate is 0.47 (zero-label ECE 0.455).

\begin{table}[t]
\centering\small
\caption{\textbf{One calibration, five domains.} PPE, 1,024 policy answers per domain. ECE before calibration (raw) and after the two Platt parameters fitted on MATH only, the fit used for RL (MATH is in-sample). ``$>$0.9'': accuracy among answers the raw readout scores above 0.9 ($n$ in parentheses). Last column: Platt refit on 100 in-domain labels, mean over 20 splits.}
\label{tab:domains}
\setlength{\tabcolsep}{4pt}
\begin{tabular}{@{}lc ccc cc c@{}}
\toprule
& & \multicolumn{3}{c}{\fx} & \multicolumn{2}{c}{Softmax judge} & \fx \\
\cmidrule(lr){3-5}\cmidrule(lr){6-7}\cmidrule(l){8-8}
Domain & Acc. & ECE raw & ECE MATH-fit & $>$0.9 & ECE raw & ECE MATH-fit & 100-label refit \\
\midrule
MATH & 0.36 & 0.084 & 0.048 & 0.87 (128) & 0.133 & 0.062 & -- \\
MMLU-Pro & 0.48 & 0.096 & 0.074 & 0.94 (224) & 0.094 & 0.115 & -- \\
IFEval & 0.50 & 0.086 & 0.075 & 0.91 (197) & 0.199 & 0.300 & -- \\
GPQA & 0.39 & 0.118 & 0.088 & 0.76 (49) & 0.100 & 0.110 & 0.075 \\
MBPP+ (code) & 0.60 & 0.162 & 0.142 & 0.74 (487) & 0.211 & 0.233 & 0.090 \\
\bottomrule
\end{tabular}
\end{table}

\paragraph{One calibration transfers across domains, except code.} The MATH fit is almost a pure shift for \fx{} ($a=1.04$), but compresses the softmax judge's scores by $a=0.72$. Carried to the other four domains without new labels, it lowers \fx{}'s ECE in every one, while it makes the softmax judge worse in all four (\cref{tab:domains}). When \fx{} gives more than 0.9, it is right 87--94\% of the time on MATH, MMLU-Pro and IFEval. The exceptions are GPQA and especially code, where answers scored above 0.9 are right only 74--76\% of the time. We read this as a limit of what the judge can know, not of the readout: a model that reads code without running it cannot see an edge case that fails a hidden test, and graduate-level science questions exceed what a 27B model knows. This is the same overconfidence we find on hard math problems (\cref{sec:hard}). It is cheap to fix where it occurs: 100 in-domain labels bring code ECE from 0.142 to 0.090 (better in 19 of 20 splits).

\begin{table}[t]
\centering\small
\caption{\textbf{Full equal-label comparison} ($N=100$ calibration labels, mean over 20 held-out splits). ``Wins'' is the number of splits in which the method's Brier beats \fx{}. AUROC applies to pointwise PPE only.}
\label{tab:e1full}
\setlength{\tabcolsep}{4.5pt}
\begin{tabular}{@{}llcccccc@{}}
\toprule
Benchmark & Method & Acc$\uparrow$ & AUROC$\uparrow$ & Brier$\downarrow$ & NLL$\downarrow$ & ECE & Wins \\
\midrule
\multirow{3}{*}{\shortstack[l]{PPE\\\scriptsize(policy answers)}} & \cellcolor{fxlight}\textbf{\fx{} decision} & \cellcolor{fxlight}\best{0.735} & \cellcolor{fxlight}\best{0.809} & \cellcolor{fxlight}\best{0.184} & \cellcolor{fxlight}\best{0.562} & \cellcolor{fxlight}0.058 & \cellcolor{fxlight}-- \\
 & Softmax (same backbone) & 0.658 & 0.709 & 0.222 & 0.642 & 0.059 & 0/20 \\
 & Scalar RM (Skywork-V2-8B) & 0.668 & 0.708 & 0.217 & 0.636 & 0.073 & 0/20 \\
\midrule
\multirow{3}{*}{JudgeBench} & \cellcolor{fxlight}\textbf{\fx{} decision} & \cellcolor{fxlight}\best{0.780} & \cellcolor{fxlight}-- & \cellcolor{fxlight}\best{0.297} & \cellcolor{fxlight}\best{0.446} & \cellcolor{fxlight}0.066 & \cellcolor{fxlight}-- \\
 & Softmax (same backbone) & 0.708 & -- & 0.373 & 0.556 & 0.067 & 0/20 \\
 & Scalar RM & 0.744 & -- & 0.352 & 0.523 & 0.063 & 1/20 \\
\midrule
\multirow{3}{*}{RM-Bench} & \fx{} decision & 0.891 & -- & 0.160 & 0.265 & 0.030 & -- \\
 & Softmax (same backbone) & 0.848 & -- & 0.219 & 0.342 & 0.036 & 0/20 \\
 & Scalar RM & \best{0.951} & -- & \best{0.079} & \best{0.144} & 0.022 & 20/20 \\
\midrule
\multirow{3}{*}{RewardBench 2} & \fx{} decision & 0.852 & -- & 0.214 & 0.427 & 0.041 & -- \\
 & Softmax (same backbone) & 0.847 & -- & 0.215 & 0.418 & 0.033 & 7/20 \\
 & Scalar RM & \best{0.894} & -- & \best{0.153} & \best{0.297} & 0.030 & 20/20 \\
\bottomrule
\end{tabular}
\end{table}

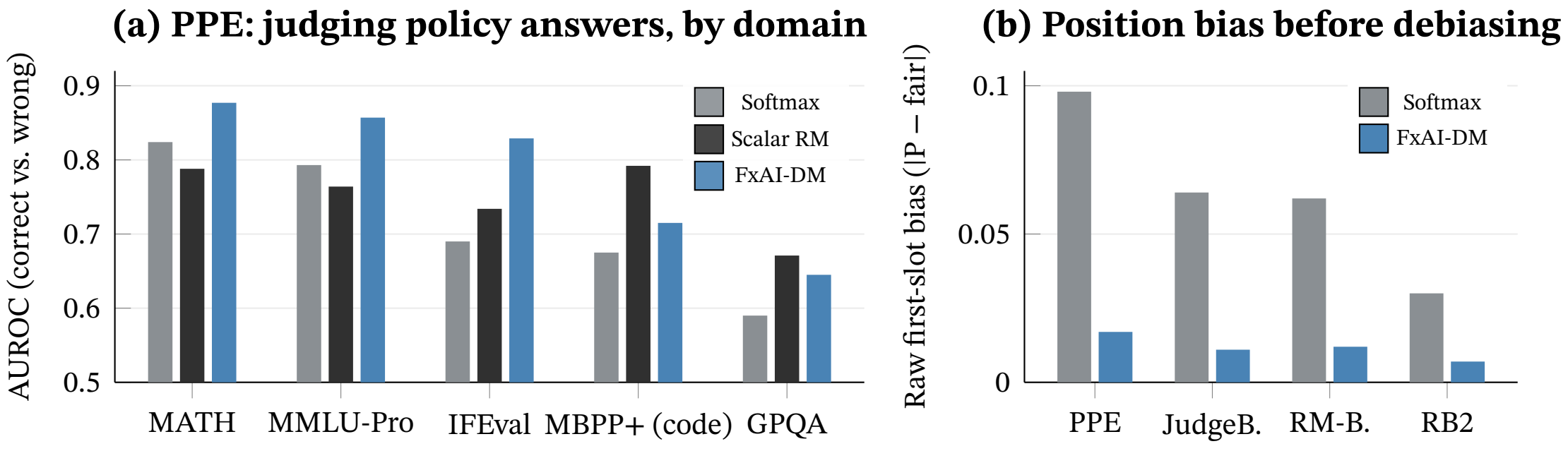
\begin{figure}[t]
\centering
\begin{tikzpicture}
\begin{groupplot}[group style={group size=2 by 1,horizontal sep=1.5cm},
  height=4.5cm, tick label style={font=\scriptsize}, label style={font=\scriptsize},
  title style={font=\small\bfseries,yshift=-3pt}, axis x line*=bottom, axis y line*=left, ymajorgrids, grid style={gray!15}]
\nextgroupplot[width=8.6cm, ybar, bar width=6.5pt, ymin=0.5, ymax=0.92,
  symbolic x coords={MATH,MMLU,IFEval,MBPP,GPQA}, xtick=data, xticklabels={MATH,MMLU-Pro,IFEval,MBPP+ (code),GPQA},
  ylabel={AUROC (correct vs.\ wrong)}, title={(a) PPE: judging policy answers, by domain}, enlarge x limits=0.13,
  yticklabel style={/pgf/number format/fixed,/pgf/number format/precision=2},
  legend style={font=\tiny,at={(0.98,0.98)},anchor=north east,draw=none,legend columns=1,fill=white,fill opacity=0.9,text opacity=1}]
\addplot[draw=none,fill=cSoft] coordinates {(MATH,0.824)(MMLU,0.793)(IFEval,0.690)(MBPP,0.675)(GPQA,0.590)};
\addlegendentry{Softmax}
\addplot[draw=none,fill=cScalar] coordinates {(MATH,0.788)(MMLU,0.764)(IFEval,0.734)(MBPP,0.792)(GPQA,0.671)};
\addlegendentry{Scalar RM}
\addplot[draw=none,fill=cOurs] coordinates {(MATH,0.877)(MMLU,0.857)(IFEval,0.829)(MBPP,0.715)(GPQA,0.645)};
\addlegendentry{\fx{}}
\nextgroupplot[width=6.2cm, ybar, bar width=9pt, ymin=0, ymax=0.105,
  symbolic x coords={PPE,JB,RMB,RB2}, xtick=data, xticklabels={PPE,JudgeB.,RM-B.,RB2},
  ylabel={Raw first-slot bias (|P $-$ fair|)}, title={(b) Position bias before debiasing}, enlarge x limits=0.2,
  yticklabel style={/pgf/number format/fixed,/pgf/number format/precision=2},
  legend style={font=\tiny,at={(0.98,0.98)},anchor=north east,draw=none}]
\addplot[draw=none,fill=cSoft] coordinates {(PPE,0.098)(JB,0.064)(RMB,0.062)(RB2,0.030)};
\addlegendentry{Softmax}
\addplot[draw=none,fill=cOurs] coordinates {(PPE,0.017)(JB,0.011)(RMB,0.012)(RB2,0.007)};
\addlegendentry{\fx{}}
\end{groupplot}
\end{tikzpicture}
\caption{\textbf{Why the decision readout is the better reward for correctness.} \textbf{(a)} It separates correct from wrong policy answers best on MATH, MMLU-Pro and IFEval; the scalar RM is better on code and GPQA, our weakest domains. \textbf{(b)} Before debiasing, the softmax readout's bias against the first option slot is 4--6$\times$ larger than the decision readout's. Without debiasing the softmax judge loses 5.4 accuracy points on PPE; the decision readout loses at most 0.8.}
\label{fig:domains}
\end{figure}

\paragraph{The advantage is in the representation.} One might ask whether the decision model wins only because its readout head is better tuned. A head-refit control (E2, \cref{app:refit}) retrains heads with the same labels on the hidden states of both models. A new head on the base Qwen3.8-27B features still stays well behind the decision-model features on PPE, JudgeBench and RM-Bench.

\subsection{Cost}
\label{sec:cost}
The decision reward needs one forward pass per question and option order, and generates no text. In E6, scoring 256 answers (2 option orders each) took 146\,s of a 340\,s training step on a separate GPU, run sequentially after generation; on hard problems the step critic for failed attempts roughly doubled step time (about 8 vs.\ 4 minutes for the answer key alone), because failed attempts are long, and the reference-free version, which asks both questions of every answer, took about 13.5 minutes per step. In E4 the decision model processed 39,896 step questions in 5,220\,s on one GPU (7.6 requests/s). A generative judge that samples 8--32 critiques of several hundred tokens per answer would need one to two orders of magnitude more decoding. Overlapping reward scoring with generation, or scoring one option order, would cut our reward time further; we have not optimized it.

\subsection{Summary: each hypothesis against its evidence}
\label{sec:verdicts}

\noindent\begin{minipage}{\linewidth}
\centering\small
\captionof{table}{\textbf{Verdicts.} The strongest piece of evidence for and against each hypothesis of \cref{tab:hyp}.}
\label{tab:verdicts}
\begin{tabularx}{\linewidth}{@{}p{0.55cm} >{\raggedright\arraybackslash}p{1.8cm} >{\raggedright\arraybackslash}X >{\raggedright\arraybackslash}X@{}}
\toprule
& Verdict & Evidence for & Evidence against / still missing \\
\midrule
H1 & \textcolor{fxteal}{\textbf{Supported}} & +4.65 MATH500 ($p<10^{-4}$); +4.0 over random reward ($p=10^{-4}$); gains grow with difficulty (\cref{fig:forest,fig:levels}) & One seed; one policy family; AIME gain +3.3 not significant alone ($p=0.09$) \\
H2 & \textcolor{fxteal}{\textbf{Supported}} on MATH / \textcolor{fxorange}{\textbf{Partly}} on hard problems & $-0.1$ vs.\ answer key on MATH500 (CI $[-1.8,1.5]$), $-0.1$ on AIME; same length change & On hard problems only, trails the answer key ($-1.2$ MATH500, $p=0.17$; lower training accuracy); parity, not superiority \\
H3 & \textcolor{fxteal}{\textbf{Supported}} & On-policy ECE 0.074 vs.\ 0.141, Brier 0.077 vs.\ 0.106; flat reward--truth gap; better ranking in all 7 group types (\cref{fig:faithful}) & Measured on math only \\
H4 & \textcolor{fxteal}{\textbf{Supported}} & ProcessBench 0.772 vs.\ 0.722 (PRM), 20/20 splits; re-solve drop 0.064 vs.\ 0.044 (softmax), random step $-0.006$ (\cref{fig:prefix,fig:pb}) & A 72B PRM would likely score higher; math only \\
H5 & \textcolor{fxteal}{\textbf{Supported}} / \textcolor{fxorange}{\textbf{Partly}} & Better Brier than same-backbone softmax on 3/4 benchmarks in 20/20 splits, tie on 1; better than scalar RM on PPE and JudgeBench; edge survives head refit (\cref{tab:e1full,tab:refit}) & Scalar RM wins RM-Bench and RewardBench 2 (preference pairs), code and GPQA \\
H6 & \textcolor{fxorange}{\textbf{Partly}} & Hybrid: best level-5 accuracy (52.1 vs.\ 50.6) and only significant pooled-AIME gain over base (+2.7, $p=0.006$) & Head-to-head vs.\ answer key not significant (+0.4); vs.\ softmax judge on the full set not significant (+1.3, $p=0.16$); one seed \\
\bottomrule
\end{tabularx}
\end{minipage}

\section{Limitations and Next Steps}
\label{sec:limits}

\begin{itemize}
  \item \textbf{One seed, one policy, math only.} The RL results use a single seed of Qwen3-1.7B on math. Our largest effects (+4 to +7 points over base and random) are well outside evaluation noise. The smaller head-to-head differences, between our reward and the softmax judge as RL rewards and between the hard-problem rewards, are unresolved and need three seeds per arm. Until then we make no policy-level claim between them.
  \item \textbf{The reference-free reward is weaker on hard problems.} Trained only on problems the policy mostly fails, it trails the answer key, and its probabilities are overconfident (mean 0.28 against a true accuracy of 0.17). Recalibrating on a hundred labelled hard answers, and training beyond 50 steps (accuracy was still rising), are the next experiments.
  \item \textbf{A large judge trains a small policy.} The reward model (27B) is much larger than the policy (1.7B). It never generates text or sees a reference answer, and the same-size softmax judge controls for size and knowledge. It cannot rule out that much of the benefit comes from a strong judge teaching a weak policy, though. Larger and non-Qwen policies are needed.
  \item \textbf{Missing RL baselines and domains.} We have not yet trained with a scalar reward model or a generative judge as the reward, or on a second domain such as instruction following (where the decision model already leads offline: IFEval AUROC 0.83 vs.\ 0.69). On preference-pair benchmarks the scalar RM is better (\cref{tab:e1full}), so fine-tuning the decision model on reward data is a natural next step.
  \item \textbf{Truncation.} With the 3,072-token limit, over 60\% of AIME answers from the two best full-set runs were cut off. Longer budgets would likely raise their AIME scores.
  \item \textbf{Context length.} \fx{} accepts at most 8,192 tokens, and we make no long-context claims.
\end{itemize}

\section{Conclusion}

A reward is only as useful as the trust one can place in its numbers. We showed that an LLM converted into a decision model, debiased and calibrated with about a hundred labels, gives a reward whose probabilities stay honest while the policy trains on it. Without ever seeing a reference answer, it improved a real model by +4.65 MATH500 points, the same as the answer key. Random-reward and same-backbone controls tie that gain to the reward. The gain grows with problem difficulty, reaching +6.9 points on the hardest MATH500 level. The same model locates reasoning errors better than a dedicated process reward model. On problems the policy mostly fails, the reference-free reward still helps but trails the answer key, because its calibration does not hold there; its step credit is then most useful added to the answer key. Calibrating for hard problems is the clearest next step. Calibrated decision rewards are a practical, cheap and auditable way to bring the success of RL beyond the tasks that come with an answer key.

\section*{Availability}

\fx{} is public under the Apache-2.0 license \citep{fxai2026dm}. All benchmarks, the policy (Qwen3-1.7B) and the training data (MATH) are public. The typed questions are given verbatim in \cref{app:prompts} and all hyperparameters in \cref{app:hparams}. Our training and evaluation code, the reward server, the hard-problem split, and every rollout, reward and per-problem evaluation result will be released at \url{https://github.com/rudransh2004/rlcdr}.

\bibliographystyle{plainnat}
\bibliography{refs}

\clearpage
\appendix
\section{Typed Questions (Verbatim)}
\label{app:prompts}

\paragraph{Outcome question (RL reward).} The state is ``Task given to an AI assistant: \{problem\} \textbackslash n\textbackslash n The assistant's response: \{response\}''. The instruction is ``Is the assistant's final answer to this math problem correct?''. The options are \emph{correct}: ``The final answer is correct'' and \emph{incorrect}: ``The final answer is wrong, missing, or not stated''. Both option orders are asked, and the reward is the calibrated mean of P(correct).

\paragraph{Critic question (step reward).} The response is split at blank lines (merged to at most 30 steps) and shown as ``Step 1: \dots''. The instruction is ``Find the earliest step that contains an error (a calculation error, a logical error, a misreading of the problem, or an unjustified or false claim). If every step is correct, choose that no step contains an error.'' The options are \emph{none}: ``No step contains an error'' and \emph{step k}: ``Step k is the first step with an error''. They are asked in forward and reversed order.

\section{Hyperparameters}
\label{app:hparams}
\begin{center}\small
\begin{tabularx}{\linewidth}{@{}lX@{}}
\toprule
Policy & Qwen3-1.7B, non-thinking chat template \\
Algorithm & GRPO, Dr.~GRPO loss, no reward standardization (TRL) \\
Prompts $\times$ samples per step & 32 $\times$ 8 \\
Steps & 100 (checkpoints every 25) \\
Learning rate & $2\times10^{-6}$, constant, no warm-up \\
KL coefficient $\beta$ & $10^{-3}$ \\
Max completion & 3,072 tokens; sampling temperature 1.0 \\
Precision & fp32 master weights, bf16 autocast, gradient checkpointing \\
Generation & vLLM (colocated), memory fraction 0.35, max length 4,096 \\
Calibration (\method{}) & Platt $a=1.036$, $b=-0.556$ (1,024 PPE-MATH items) \\
Calibration (softmax judge) & Platt $a=0.718$, $b=-0.799$ (same items) \\
Evaluation & MATH500 $n{=}4$; AIME24/25 $n{=}16$; $T{=}0.6$, top-$p$ 0.95; math-verify \\
Hardware & NVIDIA RTX PRO 6000 Blackwell (96\,GB): one GPU for the policy, one for the reward \\
\bottomrule
\end{tabularx}
\end{center}

\section{Head-Refit Control (E2)}
\label{app:refit}
One might ask whether the decision model only wins because its readout head is better tuned. We therefore retrained heads with the same labels on both models' hidden states (\cref{tab:refit}). The base Qwen3.8-27B features improve with a new head (PPE Brier 0.217 $\to$ 0.198) but stay well behind the decision-model features (0.177) on PPE, JudgeBench and RM-Bench. The two tie on RewardBench 2 (0.185 vs.\ 0.191). The shipped decision head with two Platt parameters is already within 0.01 Brier of a fully refitted head on correctness tasks, so two parameters are enough there. A refitted head does not close the gap to the scalar RM on preference benchmarks; that needs fine-tuning on preference data.

\begin{table}[h]
\centering\small
\caption{\textbf{Head refit with equal labels} (held-out Brier$\downarrow$; 1,600 labels for PPE and RM-Bench, 400 for JudgeBench and RewardBench 2).}
\label{tab:refit}
\begin{tabular}{@{}lcccc@{}}
\toprule
Features + head & PPE & JudgeBench & RM-Bench & RewardBench 2 \\
\midrule
\fx{} + Platt (shipped head, 2 parameters) & 0.179 & 0.328 & 0.157 & 0.212 \\
\fx{} head refit & 0.178 & \best{0.318} & 0.155 & 0.185 \\
\fx{} new head on features & \best{0.177} & 0.322 & 0.154 & 0.193 \\
Base Qwen3.8-27B, new head (probe judge) & 0.198 & 0.366 & 0.186 & 0.191 \\
Skywork-Reward-V2-8B + Platt & 0.210 & 0.372 & \best{0.078} & \best{0.153} \\
\bottomrule
\end{tabular}
\end{table}

\section{Full RL Results}
\label{app:full}
\begin{center}\small
\begin{tabular}{@{}lcccccccc@{}}
\toprule
& \multicolumn{4}{c}{Step 50} & \multicolumn{4}{c}{Step 100} \\
\cmidrule(lr){2-5}\cmidrule(lr){6-9}
Reward & M500 & AIME24 & AIME25 & & M500 & AIME24 & AIME25 & AIME24 p@16 \\
\midrule
Base & 72.2 & 11.3 & 9.4 & & 72.2 & 11.3 & 9.4 & 30.0 \\
Random & 71.2 & 14.0 & 8.5 & & 72.8 & 12.3 & 8.5 & 26.7 \\
Softmax & 74.8 & 12.7 & 10.0 & & 75.5 & 12.9 & 10.0 & 36.7 \\
\textbf{\method{}} & 73.9 & 14.2 & 8.8 & & \textbf{76.8} & \textbf{16.5} & 10.8 & \textbf{40.0} \\
Answer key & 74.4 & 12.9 & 13.3 & & 76.9 & 14.4 & 13.1 & 33.3 \\
\bottomrule
\end{tabular}
\end{center}
At step 50 no reward-based method is significantly different from another. A separate evaluation of the same \method{} step-50 checkpoint gave 74.85 on MATH500, which shows evaluation noise of about $\pm1$ point.

\paragraph{Hard-problem runs (E6b) at both checkpoints.}
\begin{center}\small
\begin{tabular}{@{}lccccccc@{}}
\toprule
& \multicolumn{3}{c}{Step 25} & & \multicolumn{3}{c}{Step 50} \\
\cmidrule(lr){2-4}\cmidrule(lr){6-8}
Reward & M500 & AIME24 & AIME25 & & M500 & AIME24 & AIME25 \\
\midrule
Base & 72.2 & 11.3 & 9.4 & & 72.2 & 11.3 & 9.4 \\
Answer key & 75.0 & 12.5 & 11.7 & & 75.3 & 14.8 & 9.2 \\
\fx{} only (reference-free) & 72.3 & 11.3 & 9.4 & & 74.1 & 13.1 & 10.0 \\
Answer key + \fx{} step credit & 71.9 & 11.7 & 11.7 & & 75.7 & 12.9 & 13.1 \\
\bottomrule
\end{tabular}
\end{center}
Both runs that use \fx{} start more slowly than the answer key and gain between steps 25 and 50. The reference-free step-25 checkpoint matches base on average, but its per-problem results differ from base.

\end{document}